\documentclass[journal ]{new-aiaa}
\usepackage[utf8]{inputenc}

\usepackage{textcomp}

\usepackage{graphicx}
\usepackage{amsmath}
\usepackage[version=4]{mhchem}
\usepackage{siunitx}
\usepackage{longtable,tabularx}
\usepackage[noend]{algpseudocode}
\usepackage{caption}
\usepackage{float}
\usepackage{enumitem}
\usepackage{color}
\usepackage{gensymb}
\usepackage{multirow}
\usepackage{hyperref}
\usepackage{color, colortbl}
\definecolor{Gray}{gray}{0.9}
\definecolor{LightCyan}{rgb}{0.88,1,1}

\title{Wind Tunnel Testing and Aerodynamic Characterization of a QuadPlane Uncrewed Aircraft System}
\author{Akshay Mathur\footnote{PhD Candidate, Robotics Institute, University of Michigan, AIAA Student Member}}
\affil{University of Michigan, Ann Arbor, MI, 48109}

\author{ Ella Atkins \footnote{Fred D. Durham Professor and Head, Kevin T. Crofton Aerospace and Ocean Engineering Department, Virginia Tech, AIAA Fellow} }
\affil{Virginia Tech, Blacksburg, VA 24061}

\begin{document}

\maketitle

\begin{abstract}
Electric Vertical Takeoff and Landing (eVTOL) vehicles will open new opportunities in aviation. This paper describes the design and wind tunnel analysis of an eVTOL uncrewed aircraft system (UAS) prototype with a traditional aircraft wing, tail, and puller motor along with four vertical thrust pusher motors. Vehicle design and construction are summarized. Dynamic thrust from propulsion modules is experimentally determined at different airspeeds over a large sweep of propeller angles of attack. Wind tunnel tests with the vehicle prototype cover a suite of hover, transition and cruise flight conditions. Net aerodynamic forces and moments are distinctly computed and compared for plane, quadrotor and hybrid flight modes. Coefficient-based models are developed.  
Polynomial curve fits accurately capture observed data over all test configurations. To our knowledge, the presented wind tunnel experimental analysis for a multi-mode eVTOL platform is novel. Increased drag and reduced dynamic thrust likely due to flow interactions will be important to address in future designs. 

\end{abstract}

\section*{Nomenclature}
{\renewcommand\arraystretch{1.0}
\noindent\begin{longtable*}{@{}l @{\quad=\quad} l@{}}
$C_{L},C_{D},C_{M}$ & coefficients of lift, drag and pitching moment for a given airfoil\\
$C_{SF},C_{rm},C_{ym}$ & coefficients of side force, rolling and yawing moment for a given airfoil\\
$c$ & chord length for the wing\\
$cg$ & center of gravity for the vehicle\\
$D$ & drag force acting on the vehicle\\
$L$ & lift force acting on the vehicle\\
$F_x,F_y,F_z$ & forces on the body along body axes ($x_b,y_b,z_b$)\\
$g$ & gravitational constant\\
$M_x$ & rolling moment on the vehicle\\
$M_y$ & pitching moment on the vehicle\\
$M_z$ & yawing moment on the vehicle\\
$Re$ & Reynolds number\\
$R_{wind/bf}$ & rotation matrix from wind to body frame\\
$S$ & planform area for the wing\\
$SF$ & side force acting on the vehicle\\
$T_{s_\nu}$ & static thrust force from the propulsion module as a function of electronic speed control (ESC) signal\\
$T_{s_\omega}$ & static thrust force from the propulsion module as a function of motor rotational speed\\
$T_d$ & dynamic thrust force from the propulsion module (as a function of ESC signal)\\
$T_{fwd}$ & thrust force from the forward propulsion module\\
$T_{vert}$ & total thrust force from the vertical propulsion modules\\
$V_{a}$ & magnitude of wind velocity relative to the vehicle\\
$v_{stall}$ & stall speed\\
$W$ & weight of the vehicle\\
($x_b,y_b,z_b$) & front ($x_b$), right ($y_b$) and down ($z_b$) body frame axes\\
($x_W,y_W,z_W$) & north ($x_W$), east ($y_W$) and down ($z_W$) world frame axes\\
$\alpha$ & wing/airfoil angle of attack\\
$\alpha_i$ & wing incidence angle, i.e. angle between wing chord and body X axis ($x_b$)\\ 
$\alpha_V$ & vehicle angle of attack\\
$\alpha_p$ & propeller angle of attack\\
$\alpha_{wing}$ & angle of attack for the wing\\
$\beta$ & sideslip angle\\
$\delta_a,\delta_e,\delta_r$ & angular deflections for ailerons, elevator and rudder\\
$\delta_{thr}$ & forward thrust force\\
$\gamma$ & flight path angle\\
$\phi,\theta,\psi$ & Euler angles for the body frame using Z-Y-X convention\\
$\rho$ & density of air\\
$\nu$ & ESC signal ($\micro s$)\\
$\omega$ & motor rotational speed (rpm)
\end{longtable*}}

\section{Introduction}

\lettrine{U}{rban} Air Mobility (UAM) \cite{uber_UAM_whitepaper} generalized to Advanced Air Mobility (AAM) anticipates highly maneuverable aircraft with Vertical Take-Off and Landing (VTOL) capability plus wing-based aerodynamic lift for efficient forward flight \cite{silva_vtol_2018}. Early AAM vehicle concepts featured multicopter configurations \cite{EHang,Volocopter} as multicopter designs are VTOL capable, offer excellent maneuverability, and are well studied for Uncrewed Aircraft System (UAS) applications \cite{multicopter_landslide_mapping, multicopter_use_organic_compounds_in_air, traffic_monitoring_with_multicopters}. Multicopter optimization has been studied \cite{VTOL_thrust_tilting}, but range is limited by battery energy density and inefficiency in forward flight. Conventional fixed-wing aircraft offer efficient forward flight but require a runway for take-off and landing. 
Longer range AAM vehicle concepts feature hybrid aircraft-multicopter designs such as \cite{CityAirbus,Eve,Heavyside}.  Aerodynamic performance data for AAM designs has not yet been released in the public domain.

This paper presents wind tunnel testing and aerodynamic analysis for the hybrid eVTOL QuadPlane introduced in \cite{QuadPlane_conference_paper}. 
Similar concepts combining vertical propulsion units and aerodynamic lifting surfaces to support passenger and cargo transport have been proposed  \cite{UAM_vehicles_baseline_2019}. 
In \cite{QuadPlane_conference_paper}, the authors describe a hybrid systems QuadPlane operational paradigm in which the vehicle acts as a quadrotor during taking-off, hover and landing (Quad mode), a fixed-wing aircraft during cruise (Plane mode), and a hybrid of the two during transitions between Quad and Plane modes. As a baseline concept and test platform, the QuadPlane is not aerodynamically optimal, yet it contains all pertinent elements of hybrid vehicles. It is portable, can be manufactured quickly, and offers a reconfigurable modular design. The QuadPlane was therefore determined to be an appropriate and simple baseline for wind tunnel eVTOL testing aimed at modeling multi-mode forward, vertical and transition (hybrid) performance characteristics.

An experimental procedure for static and wind tunnel testing is described for the assembled QuadPlane and its propulsion modules.
Dynamic thrust force from a propulsion module is characterized using a benchtop dynamometer inside the University of Michigan's $2' \times 2'$ wind tunnel over a wide sweep of propeller angles of attack at pertinent airspeeds.
Net forces and moments on the assembled QuadPlane are measured by a load cell inside University of Michigan's $5'\times 7'$ wind tunnel for hover (static), transition (low airspeed) and forward flight (cruise and higher airspeeds) conditions over a sequence of input commands. 
Dynamic thrust from propulsion modules is used in post-processing full-vehicle test data to distinguish airframe forces and pitching moment. 
Aerodynamic lift, drag, side-force, pitching, rolling and yawing moments are computed for each flight mode over a variety of mounting angles, tunnel speeds, and thrust conditions. Curve fit models for all aerodynamic coefficients are presented and analyzed.

This paper offers the following contributions:
\begin{enumerate}
    \item Multiple flight modes for hover, cruise and transitions are aerodynamically modeled for the same vehicle to support experimentally-validated full flight simulations. Tests conducted over a wide sweep of accelerated and trim flight conditions enable modelling steady and transition flight aerodynamics.
    \item Experimental procedure describes the use of dynamic thrust maps of the propulsion system over a wide suite of propeller angles of attack and vehicle airspeeds to achieve accurate characterization of aerodynamic forces on the vehicle.
    \item Analysis of experimental data provides insights into unconventional pitching moment and high drag forces experienced by the vehicle due to flow interactions, impacting stability and control design.  
    \item Numerical aerodynamic models for an eVTOL platform are provided for the community to use and build upon. This paper takes a step towards providing experimentally validated models for use in aircraft control and autonomy as well as next generation air traffic management including the accommodation of safe mode transitions.
\end{enumerate}

Below, Section \ref{section: background} presents a literature review of conventional and small UAS (sUAS) wind tunnel testing and a summary of published eVTOL designs. Section \ref{section: problem statement} defines paper scope and assumptions. Section \ref{section: preliminaries} presents the QuadPlane design, test apparatus and pertinent aerodynamic background for this work. The experimental procedure for this work is described in Section \ref{Section: Experimental procedure}. Results from static and wind tunnel tests for a propulsion module and assembled vehicle are shown in Section \ref{Section: Experimental results}. A discussion in Section \ref{Section: Discussion} is followed by concluding remarks in Section \ref{Section: Conclusion}.

\section{Background} \label{section: background}

This section describes related work. An overview of conventional wind tunnel testing techniques appears first, followed by a summary of literature describing recent sUAS-specific wind tunnel testing. The final section summarizes VTOL designs similar to the QuadPlane concept.

\subsection{Classical Wind Tunnel Testing}
Wind tunnels have been used for over a century to experimentally characterize and validate aircraft aerodynamic performance.  Conventional wind tunnel testing involves aerodynamic analysis of wing sections or scale models to characterize lift, drag and pitching moment at different angles of attack ($\alpha$) and Reynolds numbers ($Re$). Lift, drag and pitching moment on a wing section is characterized by coefficient-based polynomial curve fits over angle of attack and are constant for a given airfoil. Coefficients of lift, drag and pitching moment can be used to determine the lift and drag force as well as pitching moment on the wing for different dynamic pressures and wing areas. In \cite{wind_tunnel_wing_section}, the authors conduct tests to measure the lift and drag forces acting on finite wing sections with NACA 0012 airfoil, thin flat plates and thin cambered plates at low Reynolds numbers. 
Lift-Drag ratio ($L/D$), coefficient of lift ($C_L$) and coefficient of drag ($C_D$) data was plotted against $\alpha$ and the effects of Reynolds number on lift and drag was analyzed. Wind tunnel test results yield a useful conclusion that for $Re < 50,000$, thin plates are suitable as a simple NACA 0012 substitute since a curved leading edge offers little benefit at low Reynolds number.  

Full-scale aircraft have been tested as in Ref. \cite{wind_tunnel_full_aircraft} where a full scale T-tail jet was tested in a $40 \times 80$ feet subsonic wind tunnel. Stability and control characteristics were determined at angles of attack ranging from $-2\degree$ to $+42\degree$. Tests were done without components including engine nacelles, wing tip tanks and empennage. Stable trim points were determined at angles of attack between $30\degree$ and $40\degree$ depending on flap setting and center of gravity. 
Ground vehicles are also tested in wind tunnels. \cite{wind_tunnel_road_vehicle} shows an automobile on a dynamometer track undergoing wing tunnel testing. Ref. \cite{wind_tunnel_train} describes wind tunnel tests on scale models of a train, while Ref. \cite{wind_tunnel_truck_drag-reducing_devices} describes full-scale truck wind tunnel testing to improve fuel efficiency. 

\subsection{sUAS Wind Tunnel Testing}

Small UAS (sUAS) performance has also been analyzed inside wind tunnels, including multicopters. Ref. \cite{prashin's_hexacopter_wind_tunnel_journal_paper} investigates hexacopters in tractor and pusher configurations at different airspeeds and pitch angles. 
In \cite{nasa_wind_tunnel_free_flight_quad} the authors conduct free flight wind tunnel tests inside NASA's 12-foot low speed tunnel and their 20-foot vertical spin tunnel. Safety is ensured in flight by using a pyramidal tether attachment scheme. Test conditions are also simulated using models from \cite{UAS_simulation_modelling}. Vehicle pitch angles and rotor speeds are compared with simulation results for horizontal velocities of $0-60ft/s$ and for vertical descent velocities of $0-50ft/s$.  

The authors in \cite{lift_augmented_quad,wt_testing_of_LAQ,more_wt_testing_of_LAQ} present the design and aerodynamic testing of a quadcopter with short lifting wings attached at a fixed angle to the frame. The authors develop a model in \cite{lift_augmented_quad} by balancing aerodynamic forces on the vehicle and assuming power consumption varies as a cubic function airspeed. In \cite{wt_testing_of_LAQ} the authors measure lift, drag and reduction in thrust for a similar vehicle. The baseline model introduced in \cite{lift_augmented_quad} was established using wind tunnel data from the quadcopter without wings. Wind tunnel tests were conducted for different wing mounting angles and the vehicle was trimmed at each airspeed before data collection. Comparing wind tunnel data to the model showed that the model overestimated lift for most test cases indicating the need for incorporating flow interaction effects. It was shown that endurance can be improved by up to $33\%$ with lifting wings. The same platform was augmented with a blended wing-body design and a trailing edge flap in \cite{more_wt_testing_of_LAQ} and tested inside a wind tunnel with the wing mounted at $30\degree$. Lift and Drag coefficients for the vehicle with different flap deflections were plotted.  Cruise power consumption was low due to additional lift from the blended wing-body extending endurance to $140\%$ of the original quadrotor. Curve fits were plotted but no equations or aerodynamic coefficient tables were presented in these papers.

Wind tunnel tests were conducted on a scale model delta wing sUAS in \cite{wt_stryker}. The authors describe the linearized aerodynamic model and parameters obtained from wind tunnel tests. A constant airspeed of $65.62ft/s$ is used for the $2'\times2'$ wind tunnel tests for a chord based Reynolds number of $94,000$. The scale model is subjected to angles of attack ranging from $-14\degree$ to $26\degree$. The authors plot coefficients of lift, drag, side force and moments about roll, pitch and yaw axes as a function of angle of attack for different aileron and elevator commands and sideslip angles. Coefficients are shown as raw data plots assuming standard equations used for the NACA airfoil database \cite{naca_database}. Our paper presents coefficients of lift, drag, side-force, pitching, rolling and yawing moment for the QuadPlane for distinct hover, cruise and transition configurations. We establish curve fit equations for the raw data and present coefficients and errors in tabular form to facilitate follow-on  QuadPlane modeling and simulation analyses.

Ref. \cite{wt_techniques_tiltwing} describes wind tunnel testing techniques for the LA-8, a tilt wing eVTOL aircraft with eight rotors. Pros and cons of one-factor-at-a-time (OFAT) and design of experiments (DOE) approaches are shown, and a method of combining the two approaches is also explained. The authors discuss lessons learned from testing the GL-10 platform, a hybrid diesel-electric tiltwing UAS. A trim state test matrix was developed at different motor speed and dynamic pressure values through different flight stages including transition and accelerating climb out. A closed envelope was determined for each trim point wing angle for the wing angle limits at the given flight condition. The authors discuss challenges for hover testing of such platforms in a wind tunnel when the propeller flow is directed towards the floor causing flow reflection disturbances. 
Follow on work \cite{wt_prop_frame_model_ID} included a propulsion module and powered airframe testing process similar to the approach adopted in this paper. Isolated propeller tests were conducted over the range of expected operational airspeeds with propeller angle of incidence ranging from $0\degree - 180\degree$. Propulsion tests are further described in \cite{wt_prop_systemID} where thrust and torque coefficients are identified using least squares regression. Propeller incidence angles are clustered into bands with model coefficients identified for each band. The authors develop a quintic transition polynomial weighting function to smoothly move between neighboring coefficients. Ref. \cite{wt_prop_frame_model_ID} describes two modeling approaches.  The first uses only powered aircraft test data and the other uses the propeller aerodynamic model developed in \cite{wt_prop_systemID}. Propulsion forces were subtracted from the powered vehicle data to obtain the airframe aerodynamic model. The authors also discuss the possible need for supplemental propeller-airframe testing since the superposition of propulsion and airframe effects may not fully describe system behaviors.

In summary, aerodynamic data for hybrid eVTOL designs has been collected and analyzed in limited studies such as those described above, but such results are not widely documented in the literature. This paper contributes a coefficient-based wind tunnel propulsion and airframe force and moment model suite for a QuadPlane eVTOL sUAS configuration.

\subsection{eVTOL sUAS Designs and Configurations}

The Vertical Flight Society has tracked over 500 known eVTOL aircraft concepts worldwide \cite{VFS_eVTOL_directory}. NASA \cite{silva_vtol_2018} defines three categories of eVTOL concepts for AAM applications - Quadrotor, Side-by-Side Helicopter and Lift+Cruise VTOL Aircraft. The QuadPlane design falls under the Lift+Cruise category, where distinct dynamics govern each flight mode and unique control laws are required. Vehicle dynamics were simulated by the authors in \cite{QuadPlane_conference_paper} with distinct quadrotor and undisturbed fixed-wing aerodynamic and propulsion models taken from the literature.  A hybrid automaton managed control mode switching for full pattern flight.  Our new paper provides the experimental performance model suite necessary to realistically understand QuadPlane dynamics and control in simulation.

Similar designs to the QuadPlane have been proposed in the past. Ref. \cite{control_system_and_simulation_pusher_plane} shows mathematical modelling of a VTOL vehicle accompanied by stability and controllability analysis of the linearized model. Proportional-Integral-Derivative (PID) control is used in simulation results.  Ref. \cite{theoretical_autonomous_transition_nonlindyn_and_hybrid_automaton} shows a VTOL aircraft model with a linear quadratic regulator (LQR) controller design for hover and cruise, while a nonlinear controller handles transitions with a hybrid automaton. Ref. \cite{water_VTOL_flying_wing} proposes a flying wing aircraft design with a single tilting tractor propeller for an aquatic VTOL vehicle. This vehicle recharges with solar power while floating, tilts the front motor to takeoff and executes dive landing. Ref. \cite{lift_augmented_quad} describes the design and mathematical modelling of a quadrotor augmented with small lifting wings. Computational Fluid Dynamics (CFD) simulation of a lift+cruise vehicle is presented in \cite{hybrid_vtol_vehicle_aero_analysis}, while a preliminary control design is described in \cite{prelim-control-design-similar-to-quadplane}.

The QuadPlane design has high drag as a fixed wing aircraft, but it is simple and contains all the elements and operational modes of a Lift+Cruise VTOL aircraft. The QuadPlane is relatively straightforward to model. A classical quadrotor model describes hover mode, a classical fixed-wing model describes cruise mode, and a combination of the two describe transitions.  The experimental parameters presented in this paper account for the aerodynamic interactions and inefficiencies not otherwise considered in classical quadrotor or fixed-wing dynamics and control models.

\section{Problem Statement} \label{section: problem statement}

This paper experimentally characterizes the aerodynamic performance of a hybrid eVTOL UAS in quadrotor (Quad), fixed-wing airplane (Plane), and combination (Hybrid) modes. In Quad mode, the vehicle flies below fixed-wing stall speed with zero forward thrust to vertically takeoff, hover, and land. Aerodynamic control surfaces are held in neutral positions, and vehicle attitude is maintained using quadrotor control. In Plane mode, the vehicle flies as a fixed-wing aircraft, using the forward propulsion module for thrust and standard control surfaces (ailerons, elevator and rudder) for attitude control. Vertical propulsion modules are inactive in Plane mode. In Hybrid mode, typically used to transition between vertical and forward flight, all control actuators and propulsion units are active. 
Wind tunnel testing of propulsion modules and the full QuadPlane small UAS enables capture of actual performance in a manner suitable for use in follow-on modeling and simulation studies.  
This work assumes wind tunnel flow reflection is negligible and that aerodynamic model coefficients in a lookup table provide sufficient information to interpolate when switching between different flight conditions.  Aerodynamic models are assumed to have low-order polynomial curve fits.

\section{Preliminaries}\label{section: preliminaries}
This section summarizes the QuadPlane design and the wind tunnel test apparatus and process.  Coefficient based aerodynamic lift, drag, side-force, pitching, rolling and yawing moment equations are defined.

\subsection{Vehicle Design Summary}

\begin{figure}
    \centering
    \begin{tabular}{cc}
        \includegraphics[width=0.48\linewidth]{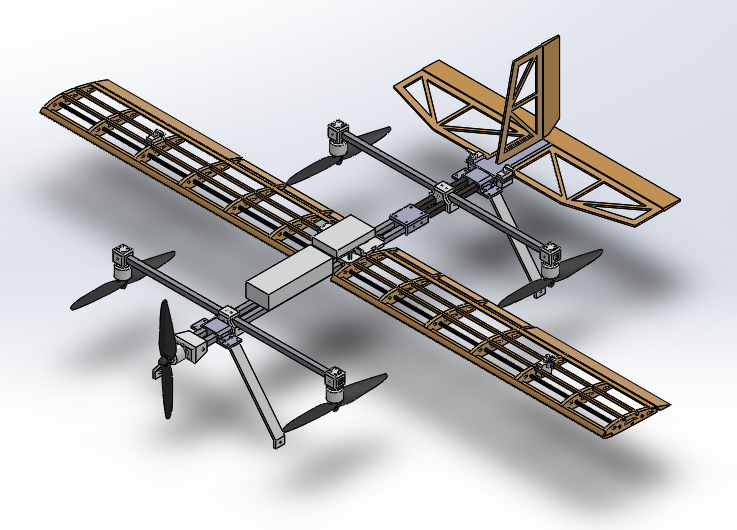} &         
        \includegraphics[width=0.46\linewidth]{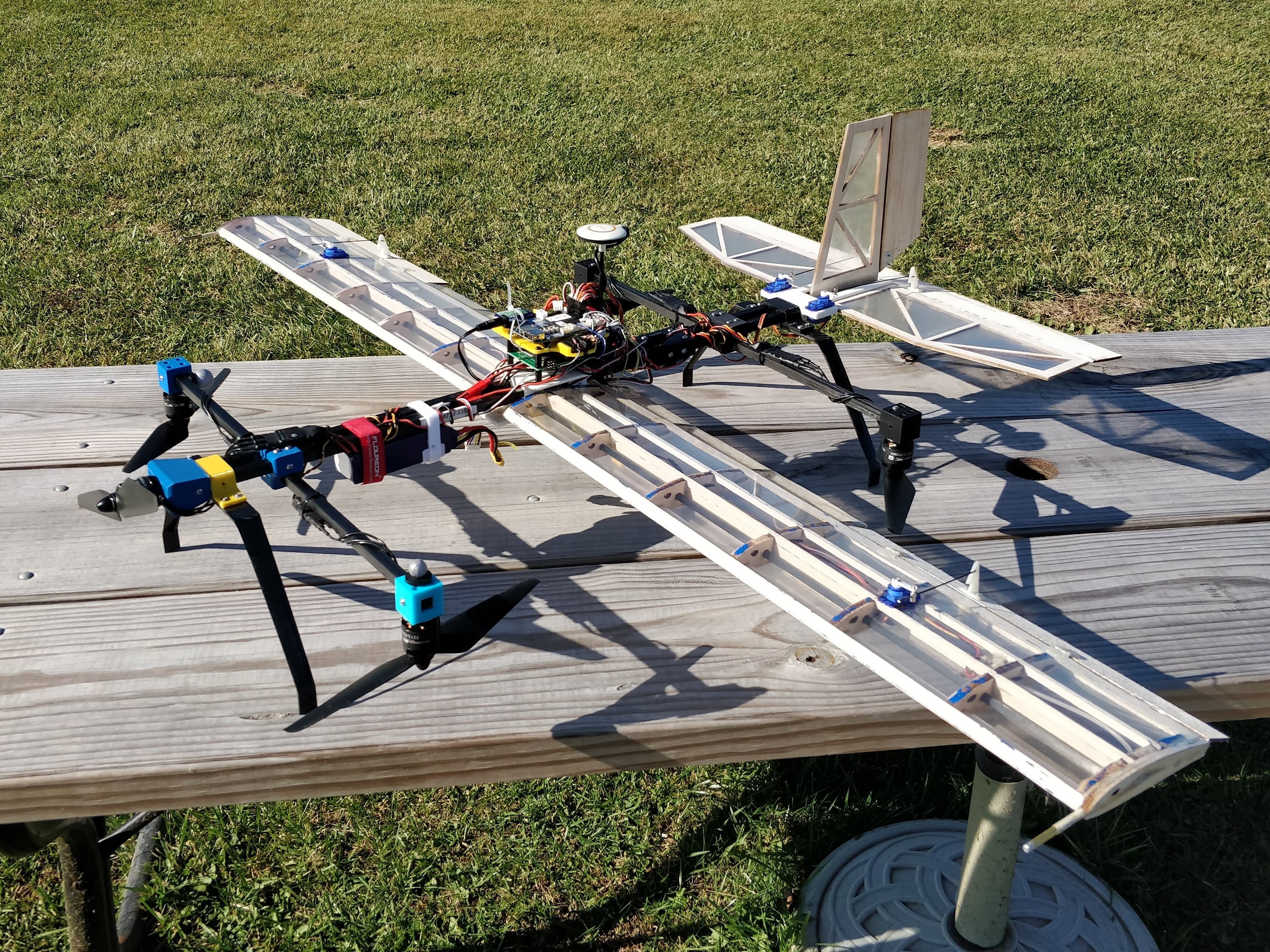}
    \end{tabular}
      \caption{QuadPlane Computer Aided Design (CAD) Schematic (left) and Balsa/Composite Prototype (right).}
      \label{fig:Design2CADandprototype}
\end{figure}

The QuadPlane was designed to fit with good clearance inside the University of Michigan's $5' \times 7'$ wind tunnel. A square pusher quadrotor layout was superimposed on a conventional tail fixed-wing airframe. Avionics and software were adapted from \cite{romano_experimental_2019,pedro's_sUAS_paper} to minimize overhead. The QuadPlane hosts four vertical thrust pusher propulsion modules, one forward thrust puller propulsion module, and servos controlling aileron, elevator and rudder deflections. The propulsion module for both forward and vertical thrust units has a $920KV$ brushless DC motor, a $30 Amp$ electronic speed controller (ESC), and a $9.5 \times 4.5$ inch propeller.  The platform is powered by a $4S$ Lithium Polymer (LiPo) battery.  The vehicle prototype is shown in Fig. \ref{fig:Design2CADandprototype}.
\begingroup
\setlength{\tabcolsep}{6pt} 
\renewcommand{\arraystretch}{1.35} 
\begin{table}[!h]
\centering
\caption{Key Vehicle Parameters}
\begin{tabular}{cc}

    \begin{tabular}{ l l l } \hline
    \multirow{2}{*}{Parameters} & \multicolumn{2}{c }{Value}\\ \cline{2-3}
    & \multicolumn{1}{l }{Imperial} & \multicolumn{1}{l }{SI Units}  \\ 
    \hline\hline
    \multicolumn{3}{ l }{Wing}\\\hline
    Span ($b$) & 48 $in$ & 1.2192 $m$ \\
    Chord ($c$) & 6 $in$ & 0.1524 $m$ \\
    Incidence angle ($\alpha_i$) & 5$\degree$ & 0.0873 $rad$\\
    Dihedral angle ($\Gamma$) & 0$\degree$ & 0 $rad$\\
    Stall speed ($v_{stall}$) & 33.10 $ft/s$ & 10.09 $m/s$\\
    Stall angle ($\alpha_{stall}$) & 15.07$\degree$ & 0.2631 $rad$\\\cline{2-3}
    Airfoil & \multicolumn{2}{c }{NACA4415}\\
    \hline\hline
    \multicolumn{3}{ l }{Stabilizers - Horizontal (hs) and Vertical (vs)}\\\hline
    Span ($b_{hs}$) & 15 $in$ & 0.3810 $m$\\
    Chord ($c_{hs}$) & 4 $in$ & 0.1016 $m$\\
    Span ($b_{vs}$) & 7 $in$& 0.1778 $m$\\
    Base chord ($c_{vs}^b$) & 4 $in$& 0.1016 $m$\\
    Tip chord ($c_{vs}^t$) & 2.5 $in$& 0.1778 $m$\\\cline{2-3}
    Airfoil  & \multicolumn{2}{c }{NACA 0006}\\
    \hline
    \end{tabular}
    & 
    \begin{tabular}{ l l l } \hline
    \multirow{2}{*}{Parameters} & \multicolumn{2}{c }{Value}\\ \cline{2-3}
    & \multicolumn{1}{l }{Imperial} & \multicolumn{1}{l }{SI Units}  \\ 
    \hline\hline
    \multicolumn{3}{ l }{Control Surfaces}\\\hline
    Aileron span ($b_{a}$) & 12 $in$& 0.3048 $m$\\
    Aileron chord ($c_{a}$) & 1.5 $in$&0.0381 $m$\\
    Elevator span ($b_{e}$) & 15 $in$& 0.3810 $m$\\
    Elevator chord ($c_{e}$) & 1.5 $in$&0.0381 $m$\\
    Rudder span ($b_{r}$) & 7 $in$& 0.1778 $m$\\
    Rudder chord ($c_{r}$) & 1.5 $in$ & 0.0381 $m$\\\cline{2-3}
    Airfoil  & \multicolumn{2}{c }{NACA 0015}\\
    \hline\hline
    \multicolumn{3}{ l }{Miscellaneous}\\\hline
    Vehicle Mass ($M$) & 3.713 $lb$ & 1.6840 $kg$\\
    Quadrotor Arm-length ($l_Q$) & 19 $in$ & 0.4826 $m$\\
    Fuselage length ($l_f$) & 35.3 $in$ & 0.8966 $m$ \\
    Propeller diameter & 9.5 $in$ & 0.2413 $m$\\
    Propeller pitch & 4.5 $in$ & 0.1143 $m$\\
    \cline{2-3}
    Battery   &  \multicolumn{2}{c }{$4S$ LiPo, 2650 $mAh$}\\
    \hline
    \end{tabular}       \\
     & 
\end{tabular}
\label{tab - vehicle params}
\end{table}
\endgroup
Table \ref{tab - vehicle params} lists pertinent design parameters for the QuadPlane. A wingspan of $4'$ was chosen based on the test section size of the wind tunnel. An aspect ratio ($AR$) of eight was chosen to support a reasonable glide ratio, yielding a wing chord of $6"$. Based on $L/D$ data for the NACA4415 wing airfoil, the wing was mounted at an angle of incidence ($\alpha_i$) of $5\degree$ relative to the fuselage to improve efficiency in forward flight. The aerodynamic center of the wing, centroid of the quadrotor motors, and vehicle center of gravity ($cg$) were co-located along the vertical axis for stability in all flight modes. The fuselage is constructed from hollow carbon fiber tubes for structural rigidity. All vehicle components are mounted on the carbon fiber tubes including propulsion modules. Mounting holes were built into the QuadPlane for secure attachment to a load cell with acceptable vibration transmission.

\subsection{Test Apparatus} \label{section: test apparatus}

\begin{figure}[!ht]
\centering
\begin{tabular}{cc}
    \includegraphics[width=0.47\linewidth]{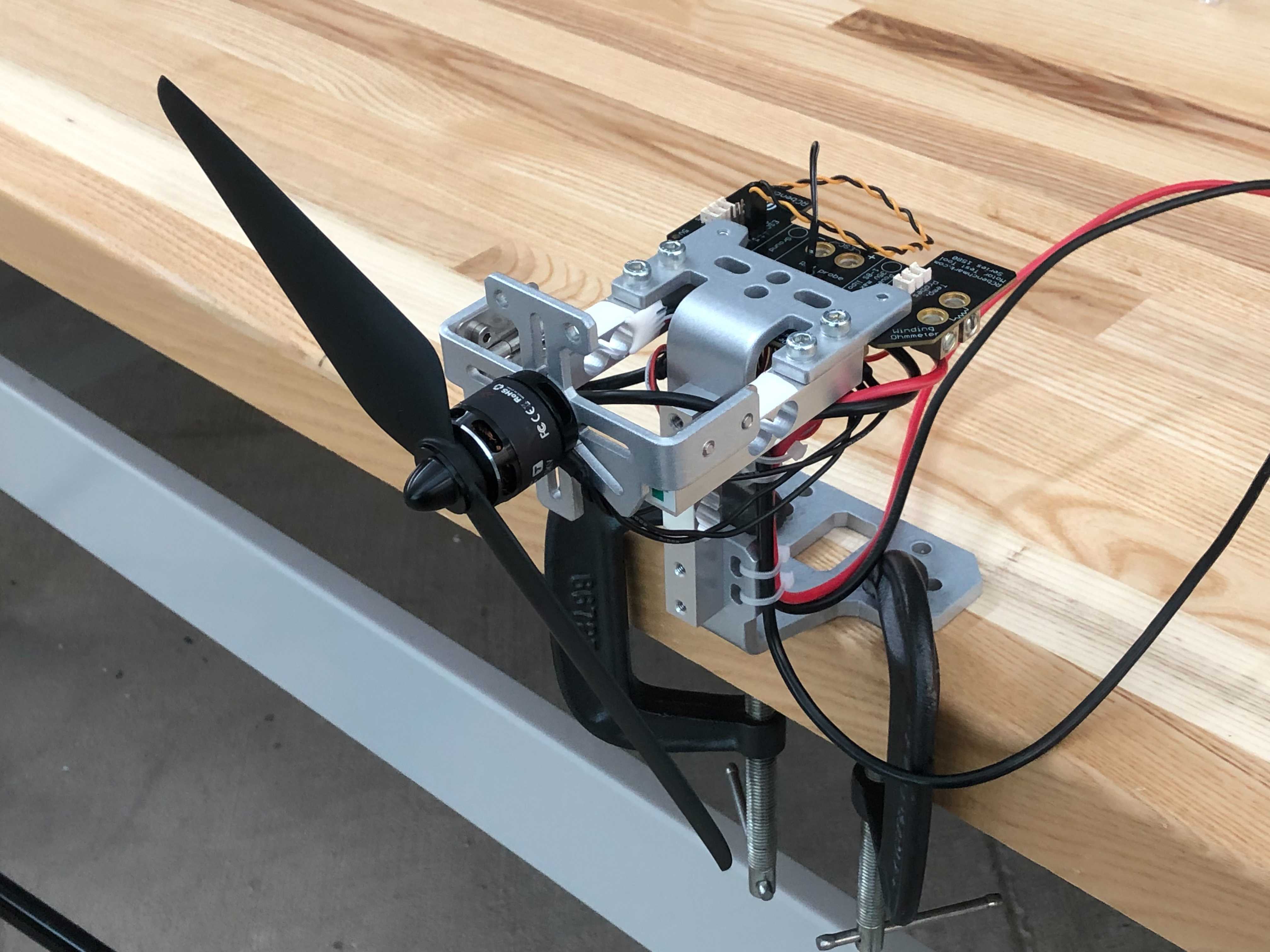}   
    &  \includegraphics[width=0.47\linewidth]{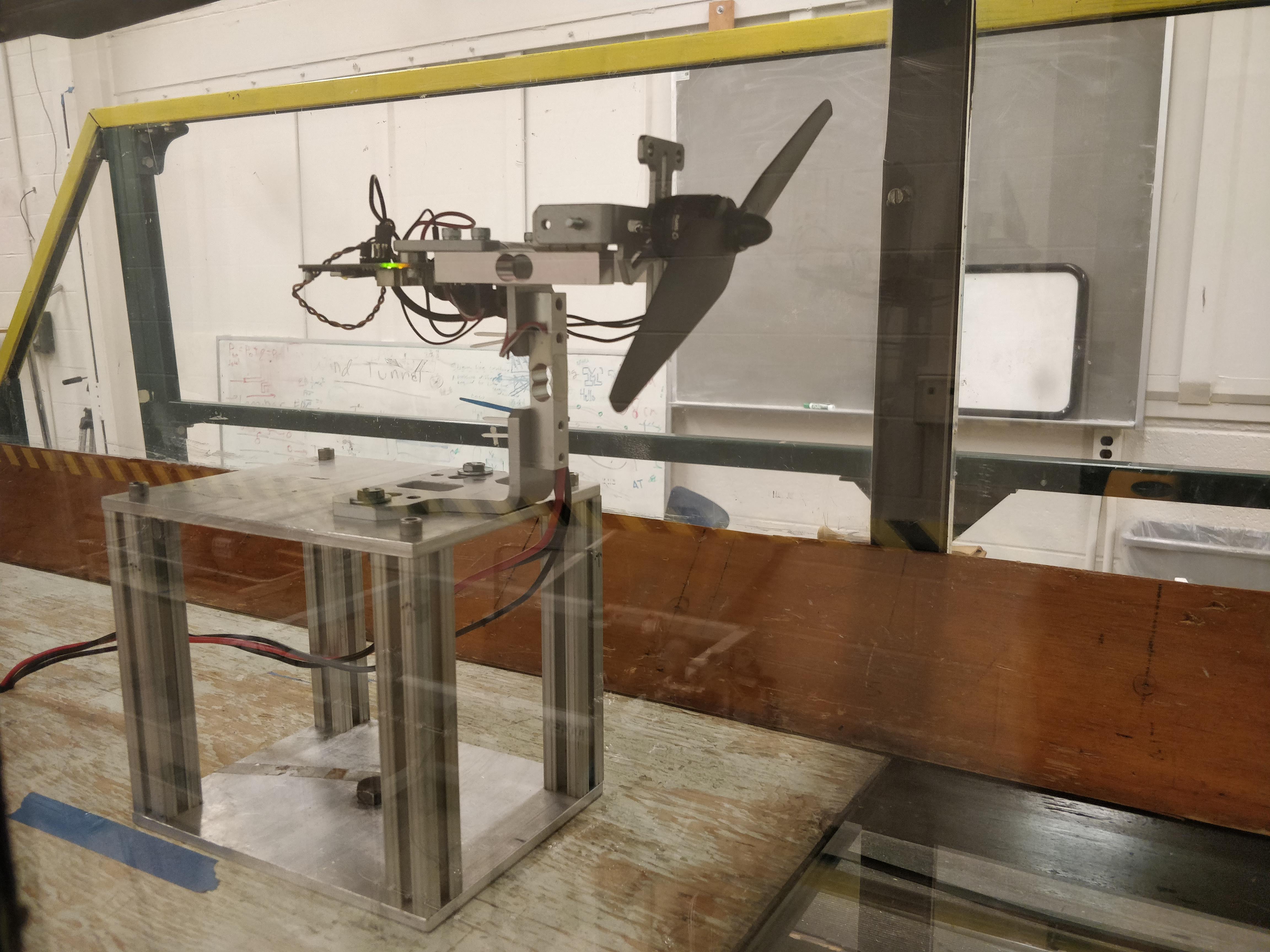}
    \end{tabular}
\caption{Dynamometer setup for static tests (left) and dynamic tests in a 2'$\times$2' wind tunnel (right).}
  \label{fig: dyno setup for static and 2x2 tests}
\end{figure}

An off-the-shelf dynamometer (RCbenchmark) was used to experimentally characterize thrust force for different conditions. The propulsion module was secured onto the dynamometer as seen in Fig. \ref{fig: dyno setup for static and 2x2 tests}. Load cell calibration must be repeated before each test series to account for environment and loading offset changes. Static thrust tests were performed in a lab.  For dynamic thrust tests, the setup was placed in a $2'\times2'$ wind tunnel test section such that the motor axis was $1'$ above the base. The setup could be fully rotated in yaw to set propeller angle of attack ($\alpha_p$). Steady state thrust, torque, rotational speed, ESC signal, current and voltage data were collected for each test case. 

\begin{figure}[!ht]
\centering
\begin{tabular}{c c}
    \includegraphics[width=0.47\linewidth]{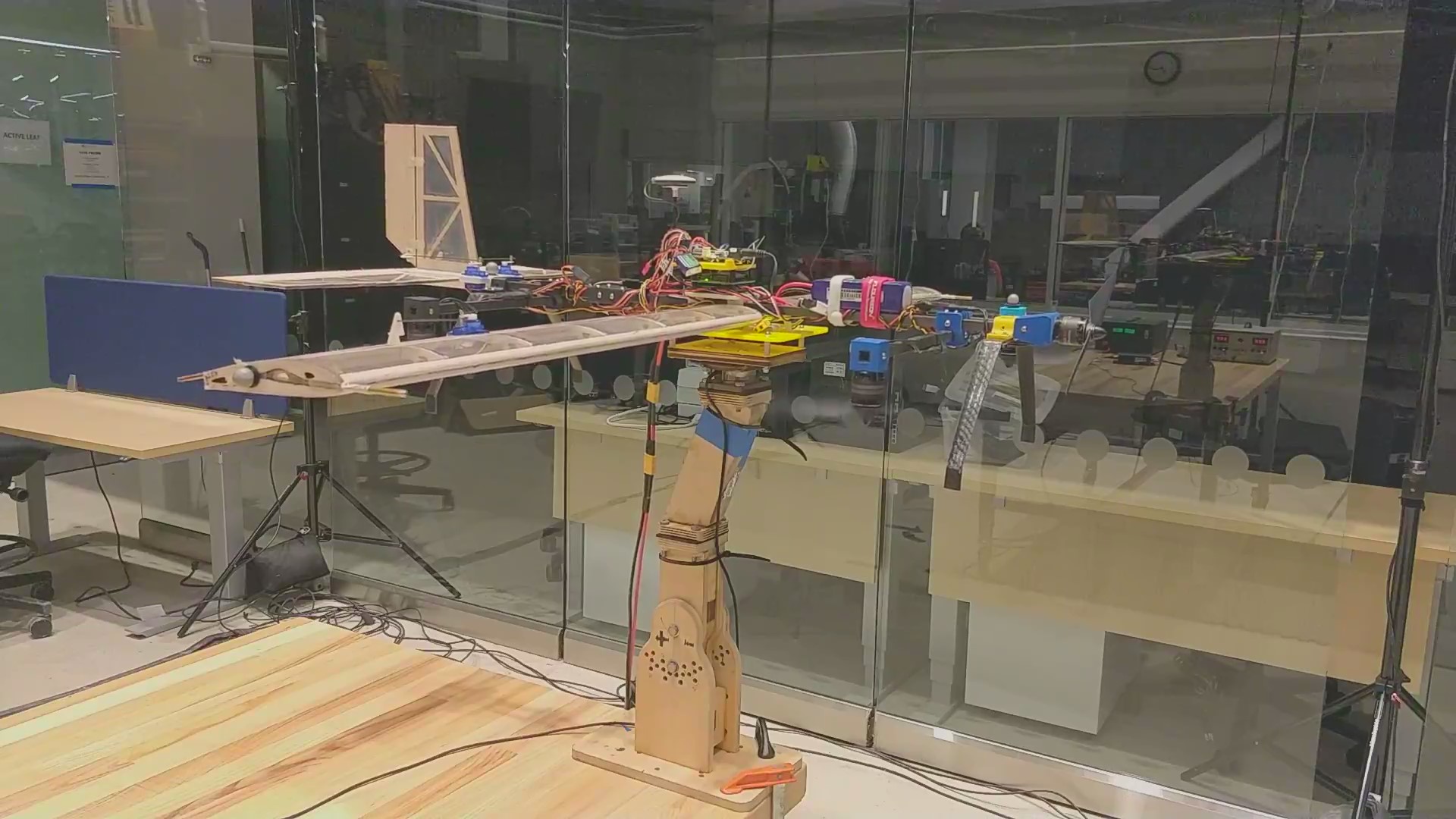} 
     & \includegraphics[width=0.47\linewidth]{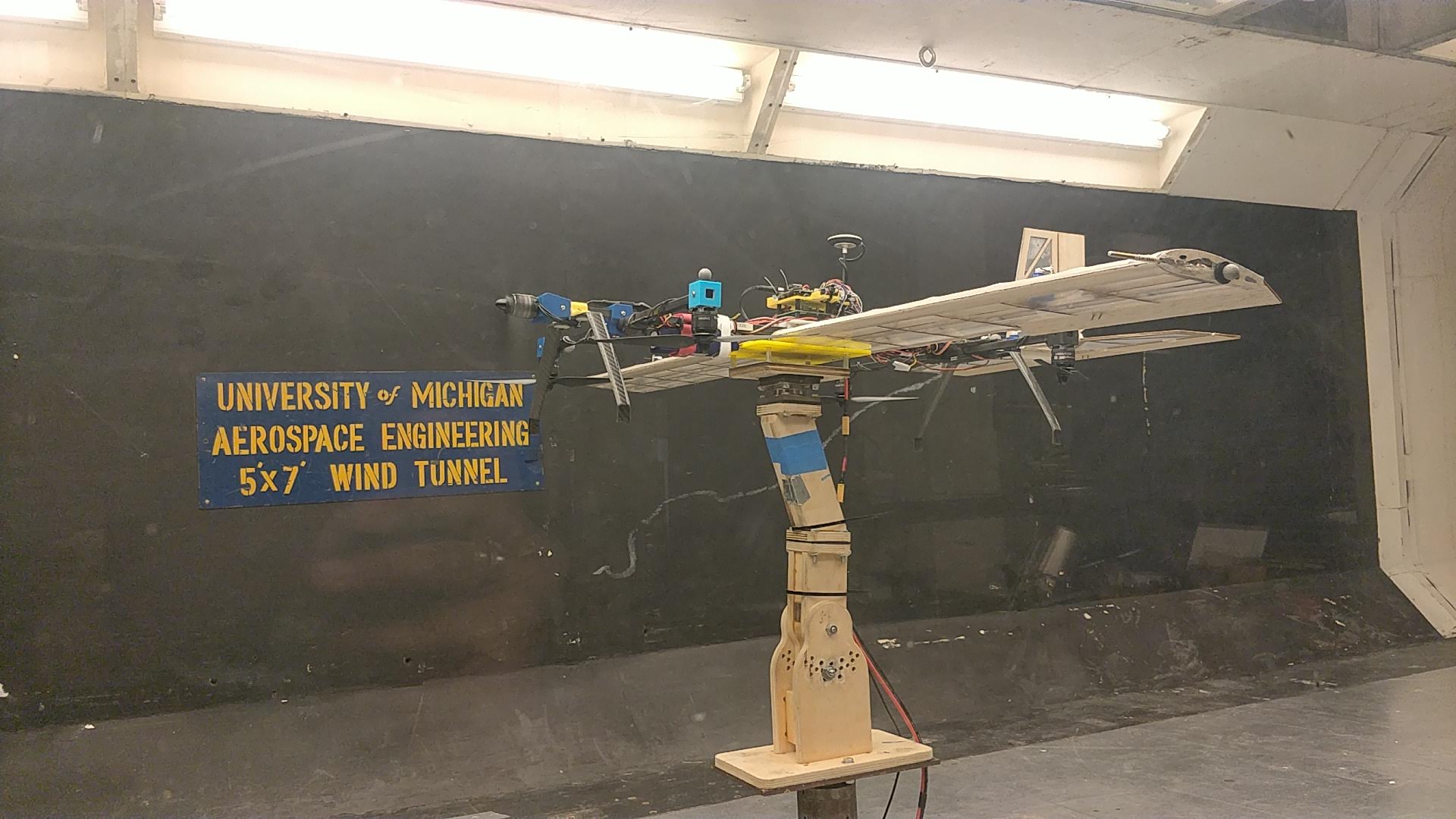}
\end{tabular}
  \caption{Assembled QuadPlane on the load cell. The left figure shows static testing in a lab. The right figure shows the apparatus inside the University of Michigan  5'$\times$7' subsonic wind tunnel.}
  \label{fig: QP in fly lab and 5x7 wind tunnel}
\end{figure}

For fully assembled testing, the QuadPlane was mounted onto a test stand equipped with a load cell as shown in Fig. \ref{fig: QP in fly lab and 5x7 wind tunnel}. The test stand could be adjusted to provide different vehicle angles of attack $\alpha_V$ in $5\degree$ increments. The load cell had strain gauges along all three axes and output voltages corresponding to the six vehicle rigid-body forces and moments. Voltages were read with 16-bit Analog to Digital Converters (ADCs) and recorded at $1000 Hz$. A calibration program from the load cell manufacturer converted voltages into forces ($N$) and moments ($Nm$). Because strain gauge output voltages vary as a function of ambient temperature, the load cell must be re-calibrated any time the temperature changed by more than $3\degree F$.  

The load cell setup was placed on a rotating platform inside the $5'\times7'$ wind tunnel to enable vehicle yaw adjustment. Both wind tunnels had instrumentation for measuring temperature as well as static and dynamic pressure inside the test section. Smoke probes inside the wind tunnels enabled flow visualization. A high-current power supply at nominal 4S battery voltage was used for all static and dynamic tests to maintain consistent power input to the system.

\subsection{Vehicle Frames and Aerodynamics} \label{section: aerodynamics}

\begin{figure}[ht]
    \centering
      \includegraphics[width=\linewidth]{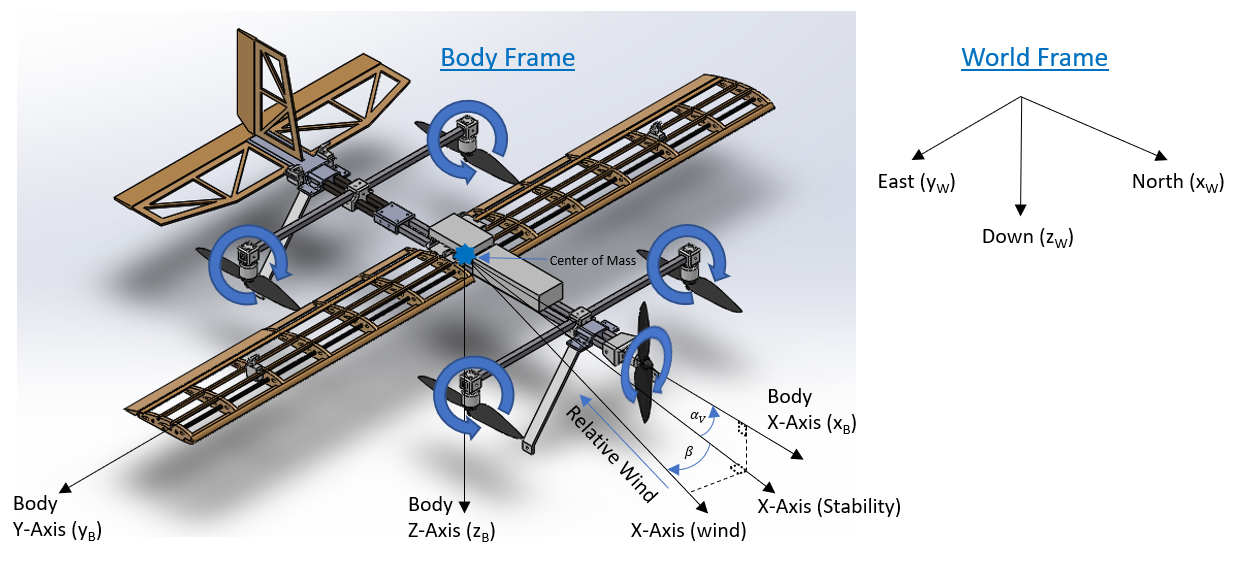}
      \caption{Coordinate frame conventions from \cite{stevens_aircraft_2015}. Curved arrows denote spin direction for each motor.}
      \label{fig: coordinate frames}
\end{figure}

Aerodynamic angle conventions and coordinate frames are consistent with reference frames from \cite{stevens_aircraft_2015}. As shown in Fig.\ref{fig: coordinate frames}, the world frame is defined by NED (North, East, Down) convention and body frame by FRD (Forward, Right, Down) convention. The wind frame $x$ axis along with the direction of rotation for each propulsion module is also shown. By convention, aerodynamic lift acts perpendicular to and drag acts along the relative wind direction. Pitching moment is defined by the right hand rule about the body Y-axis.

Airfoil lift, drag and pitching moment are dependent on flow Reynolds number per Eq. \ref{eq: Reynolds number}. At low Reynolds numbers, the flow is mostly laminar, while at high Reynolds number, flow becomes turbulent. Reynolds number is used in scaling wind tunnel scale models to full-size airframes. 

\begin{equation} 
    Re = \frac{\rho V_a l}{\mu} = \frac{\rho V_a c}{\mu}
    \label{eq: Reynolds number}
\end{equation}
where $V_a$ is vehicle airspeed; $l$ is a characteristic length such as wing chord $c$; $\rho$ denotes density; and $\mu$ is dynamic viscosity of air.      
As we wish to analyze steady level flight, flight path angle ($\gamma$) is considered zero since flow direction is always horizontal in the wind tunnel. The relationship between vehicle pitch angle ($\theta)$, vehicle angle of attack ($\alpha_V$), wing angle of attack ($\alpha_{wing}$) and flight path angle ($\gamma$) is given by Eq. \ref{eq: angular relationship}.

\begin{equation}
    \begin{aligned}
    \alpha_V & = \theta + \gamma & \implies &\alpha_V =  \theta\\
    \alpha_{wing} & = \alpha_V + \alpha_i & \implies &\alpha_{wing} = \alpha_V + 5\degree
    \end{aligned}
    \label{eq: angular relationship}
\end{equation}

\begin{figure}[!ht]
    \centering
      \includegraphics[width=\linewidth]{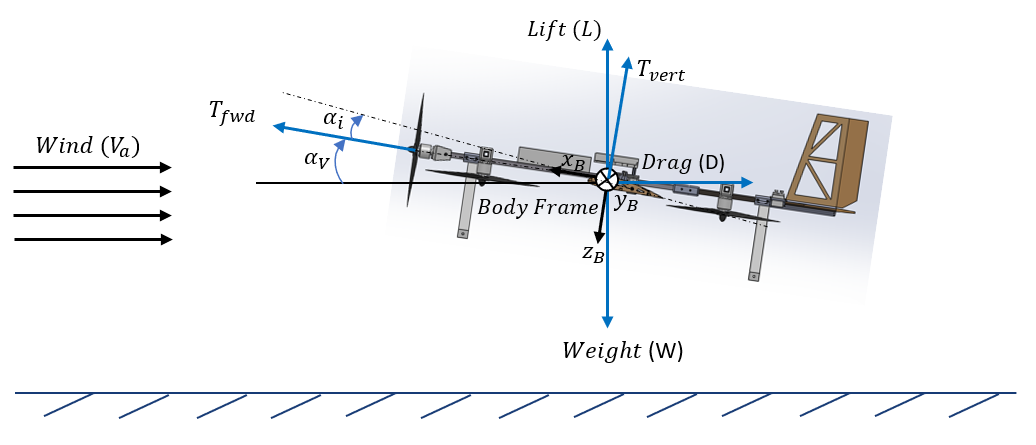}
      \caption{Free Body Diagram for the QuadPlane in level flight (Hybrid Mode). The forward propulsion module provides thrust along body $X$ and the vertical propulsion modules provide thrust along body $Z$.
      }
      \label{fig: fbd for Quadplane in steady level flight}
    \end{figure}
    
Fig. \ref{fig: fbd for Quadplane in steady level flight} shows the forces acting on the QuadPlane in Hybrid mode steady level flight with total vertical thrust ($T_{vert}$) assumed to act at the $cg$. The load cell described in Section \ref{section: test apparatus} measures net external forces and moments on the vehicle in the body frame. Balancing the forces in Fig. \ref{fig: fbd for Quadplane in steady level flight} along both body frame axes yields

\begin{equation}
    \begin{aligned}
    T_{fwd} + L.sin\theta & =  D.cos\theta + W. sin\theta\\
    W.cos\theta & = L.cos\theta + D.sin\theta + T_{vert}.
    \end{aligned}
    \label{eq: fbd for QuadPlane in steady level flight}
\end{equation}

In each test scenario, we know the forward ($T_{fwd}$) and vertical thrust ($T_{vert}$) and the vehicle weight ($W$), and we aim to find the Lift ($L$) and Drag ($D$) force on the vehicle. We model distinct "Plane" and "Quad" aerodynamic force components with simple equations that can be curve fit from wind tunnel data. We assume QuadPlane lift is primarily generated by the plane ($P$) wings, yielding the following aerodynamic lift $L_P$ and drag $D_P$ per \cite{anderson_fundamentals_2011}:

\begin{equation}
    \begin{aligned}
    L_P & = \frac{1}{2}\rho C_{L_P} V_a^2 S \\
    D_P & = \frac{1}{2}\rho C_{D_P} V_a^2 S
    \end{aligned}\label{eq:Plane Lift, Drag equations}
\end{equation}
where $C_{L_P}$ is coefficient of lift, $C_{D_P}$ is coefficient of drag, $\rho$ denotes the density of air, $V_a$ denotes airspeed and $S$ denotes wing planform area. 
$C_{L_P}$ is a linear function of angle of attack $\alpha$ until near the stall angle, when it flattens and drops as flow detaches from the wing surface. As is typical we model $C_{L_P}$ without consideration of post stall behavior.
\begin{equation}
    \begin{aligned}
    C_{L_P} & = C_{L_{P_0}} + C_{L_{P_\alpha}} \alpha
    \end{aligned}\label{eq:coeff of lift}
\end{equation}
where $C_{L_{P_0}}$ and $C_{L_{P_\alpha}}$ are dependent on Reynolds number as in the NACA tables\cite{naca_database}.
$C_{D_P}$ includes skin friction (form drag) and dynamic pressure (wing-induced drag) as given by:

\begin{equation}
    \begin{aligned}
    C_{D_P} &= C_{D_{P_f}} + C_{D_{P_i}} =  C_{D_{P_f}} + \frac{C_{L_P}^2}{\pi e AR}
\end{aligned}\label{eq: coeff of drag - Plane mode}
\end{equation}
where $C_{D_{P_f}}$ is form drag, $C_{D_{P_i}}$ is  induced drag, $e$ is Oswald efficiency factor set to $e = 0.95$ for the QuadPlane, and $AR$ is wing aspect ratio. 

QuadPlane vertical lift elements do not provide aerodynamic lift but do induce drag related linearly to airspeed while vertical thrust motors are spinning \cite{modelling_rotor_drag_effects}. Quad component  aerodynamic lift ($L_Q$) and drag ($D_Q$) are modeled as:

\begin{equation}
    \begin{aligned}
    L_Q & = 0 \\
    D_Q & = C_{D_Q} V_a = \lambda \sum_{i=1}^{4} \omega_i V_a
    \end{aligned}\label{eq:Quad Lift, Drag equations}
\end{equation}
where $C_{D_Q}$ is a lumped coefficient of drag, $\lambda$ is a positive constant, $\omega_i$ is rotational speed of the $i^{th}$ Quad motor and $V_a$ is airspeed. The overall aerodynamic lift ($L$) and drag ($D$) for the QuadPlane is given by:    
\begin{equation}
    \begin{aligned}
    L & = L_Q + L_P     &\implies L &= \frac{1}{2}\rho C_{L} V_a^2 S\\
    D & = D_Q + D_P     &\implies D & = C_{D_Q} V_a + \frac{1}{2}\rho C_{D_P} V_a^2 S
    \end{aligned}\label{eq: QuadPlane Lift, Drag equations}
\end{equation}

The vehicle experiences pitching moment due to vehicle structure, control surfaces, and differential thrust between front and rear vertical propulsion modules. We model the vehicle's pitching moment ($M_y$) using a conventional aircraft model \cite{anderson_fundamentals_2011} augmented by a differential thrust term as:
\begin{equation}
\begin{aligned}
    M_y & = \frac{1}{2}\rho c V_a^2 S C_{M} + M_{\Delta T_{vert}} 
\end{aligned}
\label{eq: pitching moment}
\end{equation}

\noindent where $c$ is a wing chord reference length, $C_M$ denotes coefficient of pitching moment incorporating the effects of changes in angle of attack and elevator deflection, and $M_{\Delta T_{vert}}$ is moment created by the differential between total front and total rear vertical thrust.
$C_M$  for an airfoil is usually modelled as a cubic function of the angle of attack. However, for a full vehicle the linear effect on pitching moment from the tail dominates, so in this work fixed-wing $C_{M}$ is modeled as a linear function of vehicle angle of attack. $C_M$ is also a linear function of the elevator deflection \cite{stevens_aircraft_2015} so $C_M$ is modelled as: 

\begin{equation}
\begin{aligned}
    C_{M} & = C_{M_{0}} + C_{M_{\alpha}} \alpha_V + C_{M_{\delta e}}\delta e 
\end{aligned}
\label{eq: coeff of pitching moment}
\end{equation}
\noindent where $C_{M_0}$, $C_{M_\alpha}$, and $C_{M_{\delta e}}$ are dependent on Reynolds number.

$M_{\Delta T_{vert}}$ depends on motor thrust commands and flow disturbances caused while flying at non-zero airspeed. As discussed previously, the front vertical thrust motors generate more thrust than the rear motors for the same input signal when $V_a\geq 0$. We model $M_{\Delta T_{vert}}$ as a quadratic function of vehicle angle of attack, even though at lower airspeeds, the quadratic term is close to zero and the behavior is quite linear.

\begin{equation}
\begin{aligned}
    M_{\Delta T_{vert}} & = M_{\Delta T_{{vert}_{0}}} + M_{\Delta T_{{vert}_{\alpha}}} \alpha_V + M_{\Delta T_{{vert}_{\alpha^2}}} \alpha_V^2 
\end{aligned}
\label{eq: pitching moment from differential thrust}
\end{equation}

In addition to the longitudinal forces and moments described above, the QuadPlane also experiences lateral moments and forces. Aileron deflections primarily induce a rolling moment ($M_x$), while also inducing a small yawing moment ($M_z$). Rudder deflection primarily induces a yawing moment and as the center of pressure for the rudder is offset from the body X axis, it also induces a rolling moment. Hence, we model rolling and pitching moments as having contributions from the ailerons and the rudder as follows:

\begin{equation}
    \begin{aligned}
        M_x &= \frac{1}{2}\rho c V_a^2 S C_{rm};  &   
        C_{rm} &= C_{rm_{\delta a}} \delta a + C_{rm_{\delta r}} \delta r \\
        M_z &= \frac{1}{2}\rho c V_a^2 S C_{ym};  &   
        C_{ym} &= C_{ym_{\delta a}} \delta a + C_{ym_{\delta r}} \delta r\\
    \end{aligned}
    \label{eq: rolling and yawing moment and coeffs}
\end{equation}
\noindent where $C_{rm_{\delta a}}$, $C_{rm_{\delta r}}$, $C_{ym_{\delta a}}$ and $C_{ym_{\delta r}}$ are the coefficients of rolling moment and yawing moment as a function of aileron and rudder deflection respectively. These coefficients are monotonic functions of the aileron and rudder deflections. 

Rudder deflections also cause a side force ($SF$) to act on the vehicle. This side force varies with the Reynold's number, like all the other forces and moments described in this paper. The side force is modelled as follows:

\begin{equation}
    \begin{aligned}
        SF &= \frac{1}{2}\rho V_a^2 S C_{SF}; & C_{SF} &= C_{SF_0} + C_{SF_{\delta r}} \delta r
    \end{aligned}
    \label{eq; side force equation}
\end{equation}
\noindent where $C_{SF}$ is the side force coefficient, which is a linear function of the rudder deflection. Ideally,  $C_{SF_0} = 0$ as the side force on the vehicle should be zero when the rudder is in its neutral position and the vehicle is experiencing zero sideslip.

All aerodynamic coefficients described here are experimentally determined from wind tunnel tests as described in the next section and are tabulated in Section \ref{Section: Experimental results}. 

\section{Experimental Procedure} \label{Section: Experimental procedure}

A suite of dynamometer and wind tunnel experiments were conducted on isolated propulsion modules and the fully assembled QuadPlane. Benchtop dynamometer tests were performed to determine static thrust profiles. Dynamic propulsion module thrust was characterized at different airspeeds ($V_a$) for a wide sweep of propeller angles of attack ($\alpha_p$) in the University of Michigan's $2' \times 2'$ wind tunnel. Forces and moments on the fully assembled vehicle were captured in the University of Michigan's $5' \times 7'$ wind tunnel with data acquired over a matrix of vehicle angles of attack ($\alpha_V$), forward and vertical thrust commands, aircraft control surface inputs and free stream airspeeds.

\subsection{Propulsion Module Tests}

\subsubsection{Static Thrust Tests}
Static thrust tests were conducted with the dynamometer-propulsion unit securely mounted in an open lab space. The test sequence issued throttle commands with pulse width modulation (PWM) values $1000 \micro s$ ($0\%$), $1250 \micro s$ ($25\%$), $1550 \micro s$ ($55\%$), $1750 \micro s$ ($75\%$) and $2000 \micro s$ ($100\%$). Steady state thrust, torque, rotational speed (rpm) and power consumption at each throttle command were measured. Tests were conducted on propulsion units in pusher and tractor configurations. Dynamometer calibration was performed at the start of each test day, a  sufficient process for the temperature-controlled lab.

\subsubsection{Dynamic Thrust Tests in $2' \times 2'$ Wind Tunnel}
Dynamic thrust for a propulsion module depends on free stream airspeed, propeller pitch and propeller angle of attack ($\alpha_p$) as well as throttle command and air temperature and pressure. To determine actual vertical and forward thrust force on the QuadPlane at different airspeeds and $\alpha_V$,  dynamic tests were performed inside Michigan's $2' \times 2'$ wind tunnel over a wide sweep of $\alpha_p$. Propeller angle of attack ($\alpha_p$) is defined as the angle between the negative of wind direction and the thrust vector, such that $\alpha_p = 0\degree$ indicates thrust  parallel to and pointing against the wind direction, such that downwash from the motor is along the ambient airflow. The dynamometer was calibrated at the start of each day of testing and whenever the temperature inside the wind tunnel changed by more than $3 \degree F$. 

The test sequence adopted the same throttle PWM set points used for static testing, but delayed data collection to allow sufficient time for the tunnel flow to stabilize at each setting. Tests were conducted at airspeeds and propeller angles of attack ($\alpha_p$) corresponding to vehicle angles of attack ($\alpha_V$) of $-5\degree$, $0\degree$, $5\degree$ and $10\degree$. For instance, $\alpha_V = 5\degree$ translates to $\alpha_p = 5\degree$ for the forward thrust module and $\alpha_p = -85\degree$ for the vertical thrust module. Table \ref{tab - 2x2 test matrix} shows the propulsion module test matrix. Reynolds number ($Re$) is calculated\footnote{http://airfoiltools.com/calculator/reynoldsnumber} using Eq.\ref{eq: Reynolds number} at $1 atm$ and $20\degree C$.

\begin{table}[h]
    \centering
    \caption{Propulsion Module Test Matrix}
    \begin{tabular}{ c c c c }
        \hline
        Wind Speed ($m/s$) & $Re$ & Propeller Angle of Attack($\alpha_p$) & Throttle \\
        \hline
        0 & 0 & 10\degree, 5\degree, 0\degree, -5\degree, 80\degree, 85\degree, 90\degree, 95\degree, 100\degree &  0\%, 25\%, 55\%, 75\%, 100\% \\ 
        5 & 50,427 & 10\degree, 5\degree, 0\degree, -5\degree, 80\degree, 85\degree, 90\degree, 95\degree, 100\degree & 0\%, 25\%, 55\%, 75\%, 100\% \\ 
        11 & 110,939 & 10\degree, 5\degree, 0\degree, -5\degree, 80\degree, 85\degree, 90\degree, 95\degree, 100\degree &  0\%, 25\%, 55\%, 75\%, 100\% \\ 
        15 & 151,281 & 10\degree, 5\degree, 0\degree, -5\degree, 80\degree, 85\degree, 90\degree, 95\degree, 100\degree & 0\%, 25\%, 55\%, 75\%, 100\% \\ 
        \hline
    \end{tabular}
    \label{tab - 2x2 test matrix}
\end{table}

\subsection{Assembled Vehicle Tests}
The fully assembled QuadPlane was tested under different laboratory simulated flight conditions. First, static tests were conducted in a large open lab area to establish a baseline data acquisition and processing pipeline and acquire zero airspeed (static) assembled vehicle forces and moments. The assembled vehicle test matrix is shown in Table \ref{tab: 5x7 test matrix}. In preparation for wind tunnel tests, the test sequence was run at each $\alpha_V$ specified in Table \ref{tab: 5x7 test matrix} under static conditions and in front of a large box fan to confirm load cell strain gauges would not saturate during wind tunnel testing.

\begin{table}[!ht]
    \centering
    \caption{Full vehicle test sequence - repeated for each combination of airspeed (0m/s, 5m/s, 11m/s, 15m/s) and $\alpha_V$ (-5$\degree$, 0$\degree$, 5$\degree$, 10$\degree$).}
    \begin{tabular}{ c c c c c c }
    \hline
    \hline
        Cumulative Time ($s$) & Quad Throttle & Fwd Throttle & Aileron & Elevator & Rudder \\
        \hline
        30 & 55\% & 0\% & 0\% & 0\% & 0\%\\
        50 & 55\% & 50\% & 0\% & 0\% & 0\% \\
        70 & 55\% & 75\% & 0\% & 0\% & 0\% \\
        90 & 55\% & 100\% & 0\% & 0\% & 0\% \\
        130 & 0\% & 75\% & 0\% & 0\% & 0\% \\
        140 & 0\% & 75\% & -80\% & 0\% & 0\% \\
        150 & 0\% & 75\% & -40\% & 0\% & 0\% \\
        160 & 0\% & 75\% & 40\% & 0\% & 0\% \\
        170 & 0\% & 75\% & 80\% & 0\% & 0\% \\
        180 & 0\% & 75\% & 0\% & -80\% & 0\% \\
        190 & 0\% & 75\% & 0\% & -40\% & 0\% \\
        200 & 0\% & 75\% & 0\% & 40\% & 0\% \\
        210 & 0\% & 75\% & 0\% & 80\% & 0\% \\
        220 & 0\% & 75\% & 0\% & 0\% & -80\% \\
        230 & 0\% & 75\% & 0\% & 0\% & -40\% \\
        240 & 0\% & 75\% & 0\% & 0\% & 40\% \\
        250 & 0\% & 75\% & 0\% & 0\% & 80\% \\
        \hline
    \end{tabular}
    
    \label{tab: 5x7 test matrix}
\end{table}

Once static tests were completed, the setup was placed at the center of the $5'\times7'$ wind tunnel test section. 
Wind tunnel tests were conducted at vehicle angles of attack ($\alpha_V$) of -5$\degree$, 0$\degree$, 5$\degree$, 10$\degree$ and airspeeds ($V_a$) of  0$m/s$ (hover), 5$m/s$ (transition), 11$m/s$ (cruise) and 15$m/s$ (faster-than-cruise).  The Table \ref{tab: 5x7 test matrix} test matrix included a baseline hover throttle condition ($55\%$) followed by increasing forward throttle to simulate transitions. The quadrotor motors were turned off and the forward throttle was set to $75\%$ for cruise (plane mode). At cruise throttle, the ailerons, elevator and rudder were deflected. Sufficient time was provided after each change in test condition to allow the airflow to settle. Multiple runs for each test condition were performed to confirm repeatability.
Collected data was post processed to derive net aerodynamic forces and moments. Dynamic thrust values from propulsion module tests were subtracted from total force values per Eq. \ref{eq: fbd for QuadPlane in steady level flight} to compute net aerodynamic body frame forces. Net forces were resolved to the wind frame to obtain aerodynamic lift and drag. Equations \ref{eq:Plane Lift, Drag equations} - \ref{eq: QuadPlane Lift, Drag equations} were then applied to compute coefficients of lift and drag. 

\section{Experimental Results} \label{Section: Experimental results}
This section presents measured data from static and wind tunnel tests. Static and dynamic thrust is characterized first. Raw QuadPlane forces and moments are presented followed by aerodynamic forces and moments and coefficients of lift and drag (Table \ref{tab: 5x7 test matrix}). Curve fitting is performed in accordance with the equations in Section \ref{section: aerodynamics}. 

\subsection{Propulsion Module Static and Dynamic Thrust}

\begin{figure}[!h]
\centering
\begin{tabular}{c c}
    \includegraphics[width=0.47\linewidth]{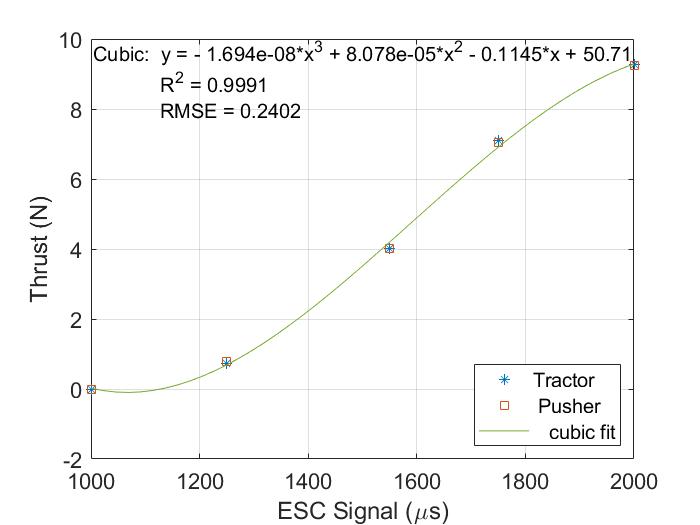}   
     &  \includegraphics[width=0.47\linewidth]{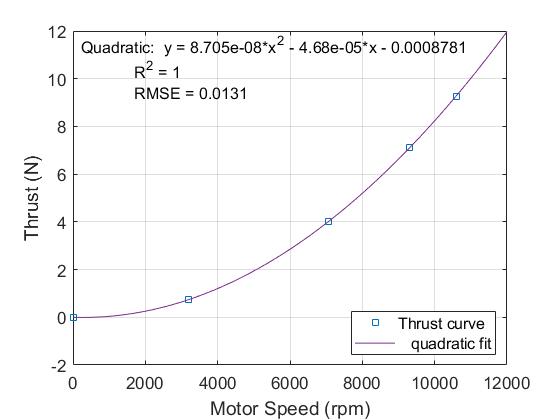}   \\
\end{tabular}
\caption{Left - Static thrust map versus input ESC signal ($\nu$) with a cubic fit. Right - Static thrust vs motor speed ($\omega$) with a quadratic fit.}
\label{fig: thrust map - pusher and tractor & thrust vs rpm}
\end{figure}

Static thrust data obtained from the dynamometer test is shown in Fig. \ref{fig: thrust map - pusher and tractor & thrust vs rpm} for both pusher and tractor configurations. Observations confirm these configurations produce nearly identical thrust without free stream or airframe influence.  As expected from the literature \cite{quan_introduction_2017}, thrust is quadratic in motor speed. Thrust is found to have a cubic relationship with ESC signal, accounting for the motor dead zone at low power inputs and impending saturation at high power inputs. Static thrust ($T_s$) curve fit equations as a function of ESC signal ($\nu$) and motor speed ($\omega$) are shown below.

\begin{equation}
    \begin{aligned}
    T_{s_\nu} = c_{\nu_0} + c_{\nu_1} \nu + c_{\nu_2} \nu^2 + c_{\nu_3} \nu^3\\
    T_{s_\omega} = c_{\omega_0} + c_{\omega_1} \omega + + c_{\omega_2} \omega^2
    \end{aligned}
    \label{eq: motor model}
\end{equation}

Dynamometer tests for the propulsion module in tractor configuration were conducted in the $2'\times2'$ wind tunnel. Fig. \ref{fig: thrust map-aoa0,aoa10,aoa80,aoa100} shows measured thrust at different airspeeds for $\alpha_p$ of $0\degree$, $10\degree$, $80\degree$ and $100\degree$. In all cases, observed thrust varies as a cubic function of the input ESC signal ($\nu$) like in static tests per Eq. \ref{eq: dynamic motor thrust model}. Thrust coefficient values are listed in Table \ref{tab: dynamic thrust coefficients}.

\begin{equation}
    T_{d} = c_{\nu_0} + c_{\nu_1} \nu + c_{\nu_2} \nu^2 + c_{\nu_3} \nu^3\\
    \label{eq: dynamic motor thrust model}
\end{equation}

\begin{figure}[!ht]
\centering
\begin{tabular}{cc}
    \includegraphics[width=0.47\linewidth]{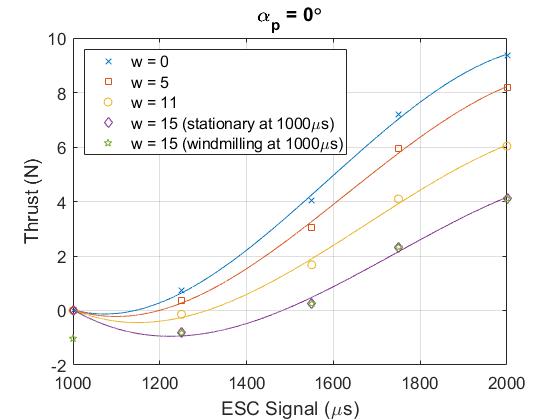} &   
    \includegraphics[width=0.47\linewidth]{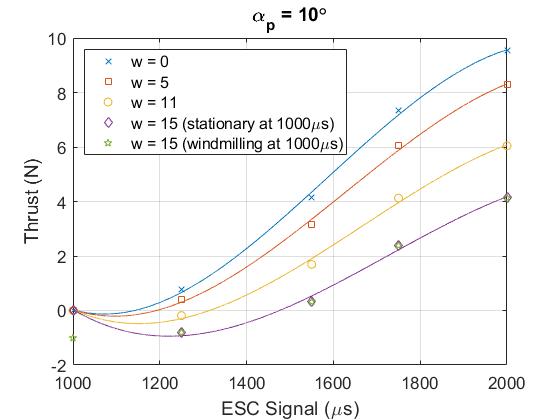}\\
    \includegraphics[width=0.47\linewidth]{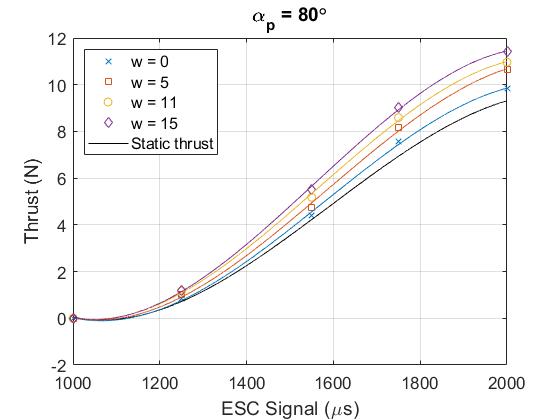} &   
    \includegraphics[width=0.47\linewidth]{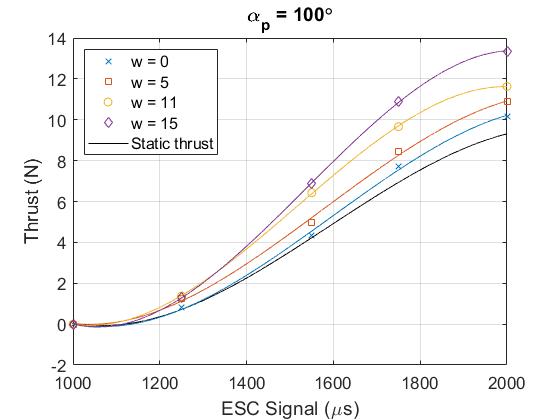}
\end{tabular}
  \caption{Dynamic thrust map for different airspeeds at different $\alpha_p$ with corresponding cubic curve fits. As airspeed increases, observed thrust decreases at lower $\alpha_p$, but increases near $\alpha_p$ of $90\degree$.}
  \label{fig: thrust map-aoa0,aoa10,aoa80,aoa100}
\end{figure}

\begin{table}[!ht]
    \centering
    \caption{Thrust coefficients for the propulsion module in dynamic conditions. Thrust curves are plotted with raw data in Fig. \ref{fig: thrust map-aoa0,aoa10,aoa80,aoa100} and Fig. \ref{fig: thrust map - wind0; wind5; w11 and w15}.}
    \begin{tabular}{ c c c c c c c c }
        \hline
        $\alpha_p$ & $V_a$ & $c_{\nu_0}$ & $c_{\nu_1}$ & $c_{\nu_2}$ & $c_{\nu_3}$ & $R^2$ & RMSE \\
        \hline
        \hline
        \multirow{4}{*}{$-5\degree$} & 0 & 54.73 & -0.1232 & 8.667e-05 & -1.820e-08 & 0.9988 & 0.2881 \\  \cline{2-8}  
        & 5 & 50.44 & -0.1111 & 7.633e-05 & -1.565e-08 & 0.9981 & 0.3142 \\ \cline{2-8}
        & 11 & 48.47 & -0.1030 & 6.826e-05 & -1.368e-08 & 0.9970 & 0.2968 \\ \cline{2-8} 
        & 15 & 47.68 & -0.0973 & 6.161e-05 & -1.194e-08 & 0.9948 & 0.2808 \\ 
        \hline
        \multirow{4}{*}{$0\degree$} & 0 & 54.43 & -0.123 & 8.627e-05 & -1.813e-08 & 0.9987 & 0.2869\\ \cline{2-8}  
        & 5 & 48.58 & -0.1069 & 7.335e-05 & -1.499e-08 & 0.9983 & 0.2915 \\  \cline{2-8} 
        & 11 & 44.04 & -0.0937 & 6.207e-05 & -1.235e-08 & 0.9968 & 0.3052 \\  \cline{2-8} 
        & 15 & 44.61 & -0.0909 & 5.725e-05 & -1.097e-08 & 0.9944 & 0.3001 \\   
        \hline
        \multirow{4}{*}{$5\degree$} & 0 & 54.11 & -0.1220 & 8.594e-05 & -1.806e-08 & 0.9990 & 0.2530\\  \cline{2-8} 
        & 5 & 50.10 & -0.1103 & 7.575e-05 & -1.551e-08 & 0.9980 & 0.3219\\  \cline{2-8} 
        & 11 & 47.41 & -0.1005 & 6.631e-05 & -1.320e-08 & 0.9971 & 0.2899 \\  \cline{2-8} 
        & 15 & 44.46 & -0.0908 & 5.736e-05 & -1.103e-08 & 0.9928 & 0.3416 \\ 
        \hline
        \multirow{4}{*}{$10\degree$} & 0 & 55.58 & -0.1253 & 8.829e-05 & -1.858e-08 & 0.9990 & 0.2595\\  \cline{2-8} 
        & 5 & 48.96 & -0.1080 & 7.431e-05 & -1.523e-08 & 0.9984 & 0.2893\\  \cline{2-8} 
        & 11 & 46.45 & -0.0989 & 6.564e-05 & -1.314e-08 & 0.9970 & 0.2945\\  \cline{2-8} 
        & 15 & 45.96 & -0.0940 & 5.960e-05 & -1.152e-08 & 0.9943 & 0.3064\\ 
        \hline
        \multirow{4}{*}{$80\degree$} & 0 & 55.49 & -0.1257 & 8.902e-05 & -1.879e-08 & 0.9993 & 0.2296\\  \cline{2-8} 
        & 5 & 55.95 & -0.1272 & 9.031e-05 & -1.901e-08 & 0.9990 & 0.2840\\  \cline{2-8} 
        & 11 & 57.55 & -0.1322 & 9.491e-05 & -2.022e-08 & 0.9994 & 0.2249\\  \cline{2-8} 
        & 15 & 62.32 & -0.1434 & 1.032e-04 & -2.210e-08 & 0.9997 & 0.1746\\ 
        \hline
        \multirow{4}{*}{$85\degree$} & 0 & 56.91 & -0.1278 & 8.970e-05 & -1.873e-08 & 0.9980 & 0.3924 \\  \cline{2-8} 
        & 5 & 55.19 & -0.1255 & 8.905e-05 & -1.87e-08 & 0.9989 & 0.3047\\  \cline{2-8} 
        & 11 & 55.47 & -0.1289 & 9.365e-05 & -2.016e-08 & 0.9998 & 0.1427\\  \cline{2-8} 
        & 15 & 58.03 & -0.1362 & 9.986e-05 & -2.167e-08 & 1.0000 & 0.0679\\ 
        \hline
        \multirow{4}{*}{$90\degree$} & 0 & 46.57 & -0.1056 & 7.435e-05 & -1.533e-08 & 0.9999 & 0.1011 \\  \cline{2-8} 
        & 5 & 52.76 & -0.1204 & 8.565e-05 & -1.797e-08 & 0.9989 & 0.3095\\  \cline{2-8} 
        & 11 & 61.10 & -0.1420 & 1.034e-04 & -2.246e-08 & 1.0000 & 0.0253\\  \cline{2-8} 
        & 15 & 68.55 & -0.1600 & 1.170e-04 & -2.557e-08 & 1.0000 & 0.0315\\ 
        \hline
        \multirow{4}{*}{$95\degree$} & 0 & 59.39 & -0.1334 & 9.364e-05 & 1.960e-08 & 0.9971 & 0.4781 \\  \cline{2-8} 
        & 5 & 53.20 & -0.1218 & 8.698e-05 & -1.834e-08 & 0.9989 & 0.2969\\  \cline{2-8} 
        & 11 & 64.82 & -0.1509 & 1.101e-04 & -2.403e-08 & 1.0000 & 0.0327\\  \cline{2-8} 
        & 15 & 72.45 & -0.1695 & 1.244e-04 & -2.727e-08 & 0.9999 & 0.0895\\ 
        \hline
        \multirow{4}{*}{$100\degree$} & 0 & 55.82 & -0.1257 & 8.836e-05 & -1.846e-08 & 0.9989 & 0.2892\\ \cline{2-8}  
        & 5 & 51.63 & -0.1189 & 8.537e-05 & -1.804e-08 & 0.9988 & 0.3248\\  \cline{2-8} 
        & 11 & 70.86 & -0.1653 & 1.210e-04 & -2.659e-08 & 0.9999 & 0.0731\\  \cline{2-8} 
        & 15 & 85.51 & -0.1964 & 1.416e-04 & -3.073e-08 & 1.0000 & 0.0751\\
        \hline
    \end{tabular}
    
    \label{tab: dynamic thrust coefficients}
\end{table}

At lower $\alpha_p$ values, dynamic thrust for a given $\nu$ decreases as airspeed increases relative to static thrust. For instance, at $\alpha_p$ of $0\degree$, full throttle thrust at airspeed $15m/s$ is only $44\%$ of its static counterpart. This occurs because the spinning propeller disc experiences higher drag force at higher free stream flow conditions. Negative thrust values at $25\%$ throttle ($1250\micro s$) for $11m/s$ and $15m/s$ indicate drag is higher than generated thrust. At higher airspeeds ($V_a = 15m/s$), the propeller keeps spinning at the end of the test sequence due to windmilling. A corresponding negative thrust value for the second test run is observed and plotted as a five-point star for $\alpha_p = 0\degree$ at $\nu = 1000\micro s$. At the maximum test airspeed ($15m/s$), a net positive thrust from the propeller is observed only at or above $50\%$ throttle. 

At higher $\alpha_p$, e.g., near $\alpha_p = 90\degree$, the observed trend is reversed. For a given $\nu$, thrust increases at higher airspeeds as shown in plots for $\alpha_p$ of $80\degree$ and $100\degree$ in Fig. \ref{fig: thrust map-aoa0,aoa10,aoa80,aoa100}. This is due to a combination of two factors. First, due to the $2'$ width of the test section, downwash from the propeller reflects from the wind tunnel wall resulting in an increase in thrust compared to unrestricted flow conditions. We observe thrust measured at $V_a = 0m/s$ for all high magnitude $\alpha_p$ values is higher than static thrust in Fig. \ref{fig: thrust map - pusher and tractor & thrust vs rpm}. Second, airflow perpendicular to the motor axis reduces downwash from the propeller and causes a pseudo ground effect. This phenomenon is described in \cite{wt_techniques_tiltwing} and summarized above in \ref{section: background}.

\begin{figure}[!ht]
\centering
\begin{tabular}{cc}
    \includegraphics[width=0.47\linewidth]{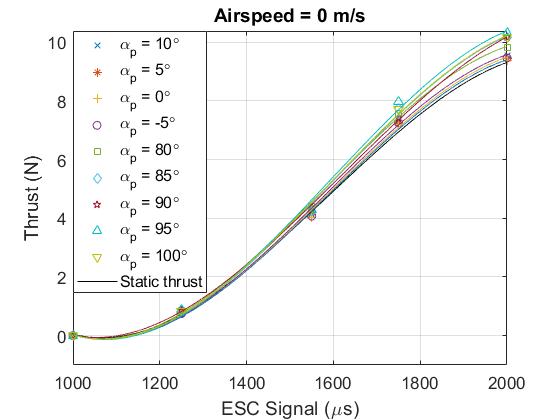} &   
    \includegraphics[width=0.47\linewidth]{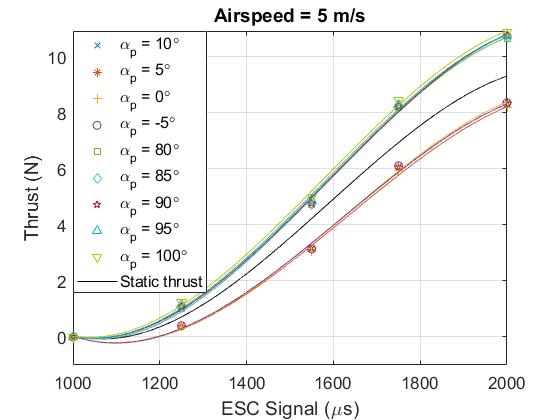}\\
    \includegraphics[width=0.47\linewidth]{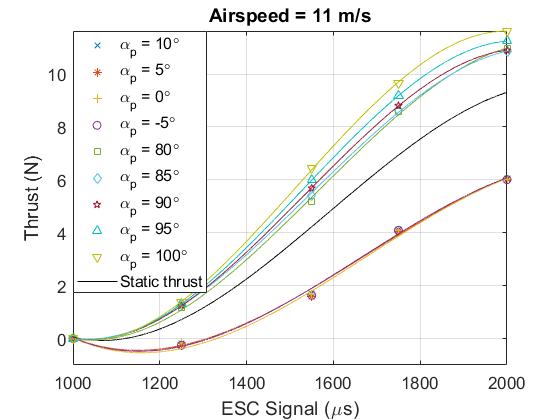} &   
    \includegraphics[width=0.47\linewidth]{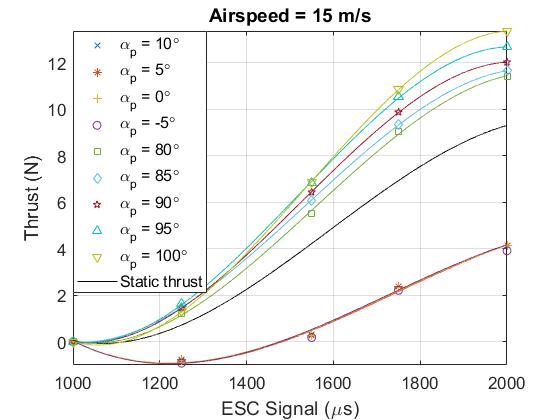}
\end{tabular}
  \caption{Thrust map for different $\alpha_p$ and airspeeds. }
  \label{fig: thrust map - wind0; wind5; w11 and w15}
\end{figure}

Fig. \ref{fig: thrust map - wind0; wind5; w11 and w15} shows thrust for different $\alpha_p$ at a given airspeed ($V_a$). As a baseline, tunnel measurements at $V_a = 0m/s$ and $\alpha_p$ = $0\degree$ match unobstructed static thrust test data as shown in Fig. \ref{fig: thrust map - pusher and tractor & thrust vs rpm}. Intuitively, static tests in an unobstructed volume would show all curves are independent of $\alpha_p$, but wind tunnel flow reflection causes slight increase in thrust data, especially for higher $\alpha_p$. At low $V_a$ values, thrust magnitude is not strongly dependent on $\alpha_p$. At  higher $V_a$, we see a clear separation of curves with dynamic thrust increasing near $\alpha_p \sim 90\degree$. For $\alpha_p > 90\degree$, thrust is measured along the airflow so propeller drag appears to actually increase measured thrust. The highest thrust is measured at $V_a = 15m/s$ and $\alpha_p = 100\degree$ at full throttle.

The largest $\alpha_p$ values ($80\degree - 100\degree$) are experienced by the vertical pusher propulsion modules in forward flight. The pusher propulsion module propellers were much too close to tunnel walls, so we use experimental tractor configuration thrust models to represent vertical thrust in this paper. Given our $\alpha_p$ convention and nearly identical thrust values from pusher and tractor configurations per Fig. \ref{fig: thrust map - pusher and tractor & thrust vs rpm}, dynamic thrust is assumed the same from a pusher or tractor propeller at a given $\alpha_p$. Dynamic thrust results from individual propulsion units are used for further analysis in Section \ref{section: full vehicle windtunnel tests}.

\subsection{Assembled Vehicle Wind Tunnel Tests} \label{section: full vehicle windtunnel tests}

\begin{figure}[!ht]
\centering
\includegraphics[width=0.7\linewidth]{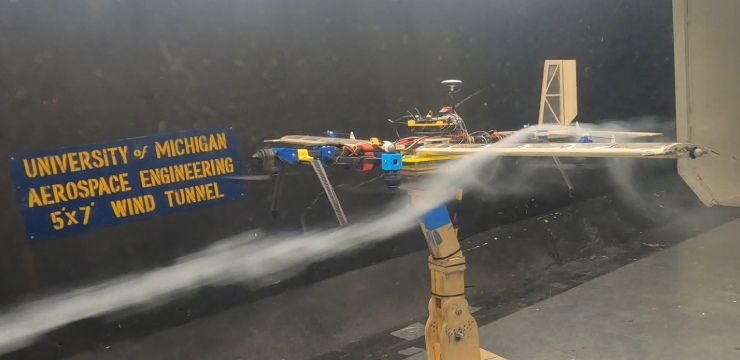}
  \caption{Flow around the QuadPlane during transition flight conditions.}
  \label{fig: flow visualization in trasition}
\end{figure}

\begin{figure}[!ht]
\centering
\begin{tabular}{cc}
    \includegraphics[width=0.47\linewidth]{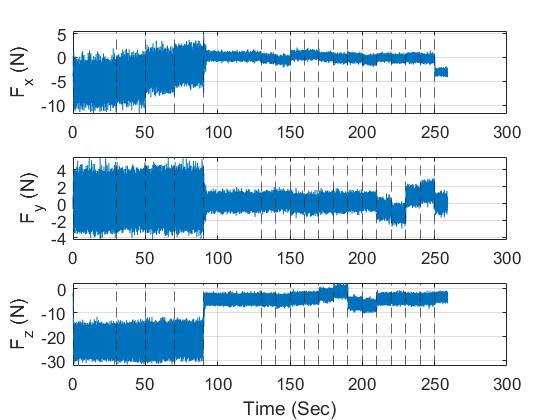}
     & \includegraphics[width=0.47\linewidth]{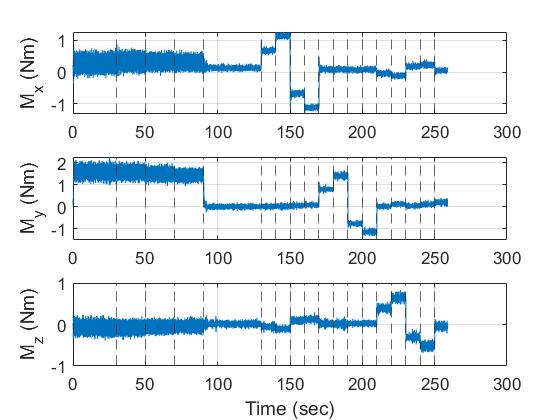}
\end{tabular}
  \caption{Forces (left) and Moments (right) on the QuadPlane with $\alpha_V = 0\degree$ at 11m/s airspeed. Vertical lines denote a change in throttle command per Table \ref{tab: 5x7 test matrix}.}
  \label{fig: forces and moments - 11m/s aoa0}
\end{figure}

Fig. \ref{fig: flow visualization in trasition} shows the assembled QuadPlane undergoing tests in the University of Michigan's $5'\times 7'$ wind tunnel with smoke  illustrating flow interaction between the left two vertical propulsion modules and the wing during transition flight, i.e. all five motors active.  No flutter was observed during tests indicating a structurally sound airframe. Raw forces and moments at $\alpha_V = 0\degree$ and $V_a = 11m/s$ are shown in Fig. \ref{fig: forces and moments - 11m/s aoa0}. The black vertical lines denote a change in input signal per Table \ref{tab: 5x7 test matrix}. The load cell was calibrated in static conditions, so all forces and moments seen after the end of the $250 s$ data collection sequence are nominal values for the airframe at the given airspeed. Control surface deflections provide higher roll ($M_x$), pitch ($M_y$) and yaw ($M_z$) moments than required for normal flight conditions at this airspeed. We observe small forces on the vehicle from elevator and rudder deflections as expected. 

When the vertical motors are spinning a large net positive pitching moment is observed. This pitching moment is caused by the rear two vertical propulsion modules producing less vertical thrust than the front modules due to upstream disturbances in airflow caused by downwash from the front motors and vehicle structure.  For the test case in Fig.\ref{fig: forces and moments - 11m/s aoa0} ($\alpha_V$ = $0\degree$, airspeed = $11m/s$), we see a net pitching moment of $1.56Nm$ about the center of gravity ($cg$) equivalent to a force of $3.12 N$ acting at a distance of $l_Q$ from the $cg$. The baseline no-thrust pitching moment in Fig. \ref{fig: forces and moments - 11m/s aoa0} is close to $0 Nm$, indicating the airframe introduces negligible pitching moment at this airspeed. Dynamic thrust tests show at this airspeed and $\alpha_V$ a single propulsion module produces $5.69N$ of thrust. Rear propulsion modules generate $27.5\%$ lower thrust than front modules for this test case. 

The load cell measures forces ($F_x$,$F_y$,$F_z$) in the body frame. To determine lift $L_{exp}$ and drag $D_{exp}$  for each test case, we determine the net aerodynamic forces in the body frame and then transform them into the wind frame. Net aerodynamic force is computed by subtracting dynamic thrust for given $\alpha_V$, $V_a$ and $\nu$ values from $F_x,F_y,F_z$ per Eq. \ref{eq: experimental lift drag}. Note that $\alpha_p = \alpha_V$ for the forward thrust propulsion module and $\alpha_p = \alpha_V + 90\degree$ for the vertical thrust propulsion modules.

\begin{equation}
    \begin{aligned}
    L_{exp} = -(F_z + T_{vert}).cos\alpha_V + (F_x - T_{fwd}).sin\alpha_V \\
    D_{exp} = -(F_z + T_{vert}).sin\alpha_V - (F_x - T_{fwd}).cos\alpha_V    
    \end{aligned}
    \label{eq: experimental lift drag}
\end{equation}
\noindent where $T_{vert}$ and $T_{fwd}$ are vertical and forward thrust, respectively. Thrust from the front two motors ($T_{vert_{fr}}$) is considered to be the same as dynamic thrust observed in the isolated propulsion module test. Thrust from the rear two motors ($T_{vert_{rr}}$) is calculated from the observed pitching moment for the test case as explained in the previous paragraph using the equation below:  

\begin{equation}
    \begin{aligned}
    T_{vert} & = 2. T_{vert_{fr}} + 2. T_{vert_{rr}} & = 2. T_{d}  + 2. (T_d - M_y/l_Q)
    \end{aligned}
    \label{eq: dynamic vertical trust compensated my pitching moment}
\end{equation}

\begin{figure}[!ht]
    \centering
    \begin{tabular}{cc}
        \includegraphics[width=0.47\linewidth]{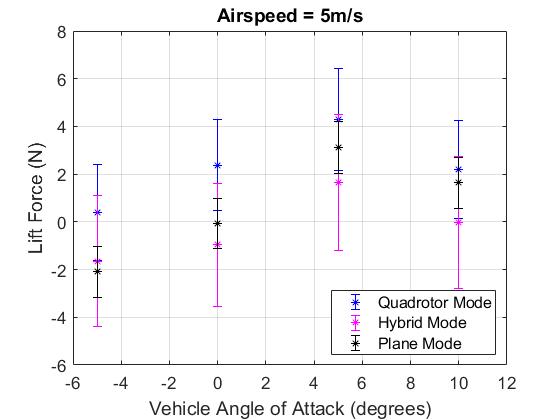}
        & \includegraphics[width=0.47\linewidth]{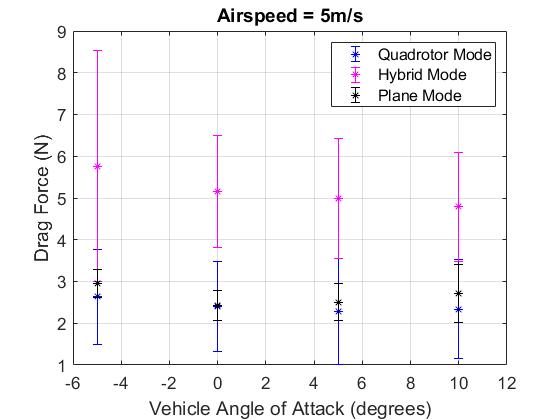}\\
        \includegraphics[width=0.47\linewidth]{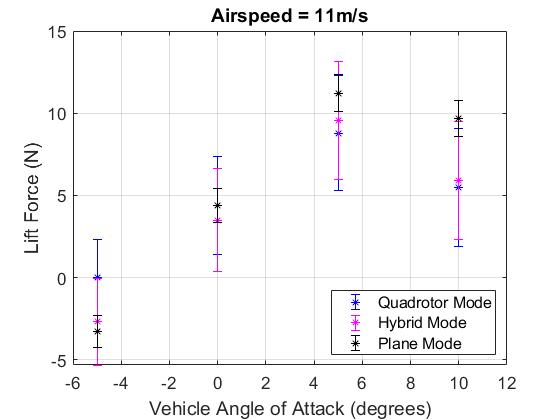}
        & \includegraphics[width=0.47\linewidth]{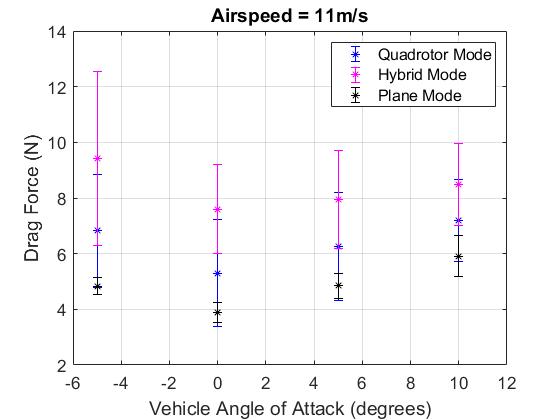}\\
        \includegraphics[width=0.47\linewidth]{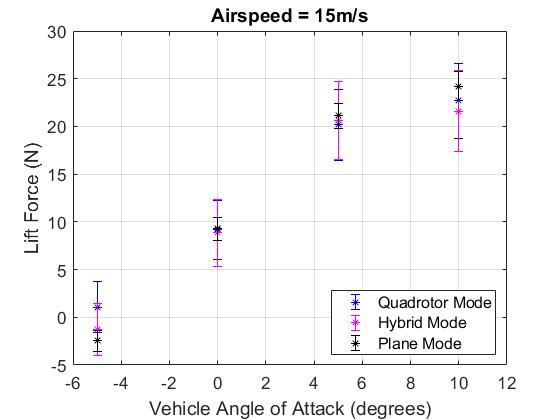}
        & \includegraphics[width=0.47\linewidth]{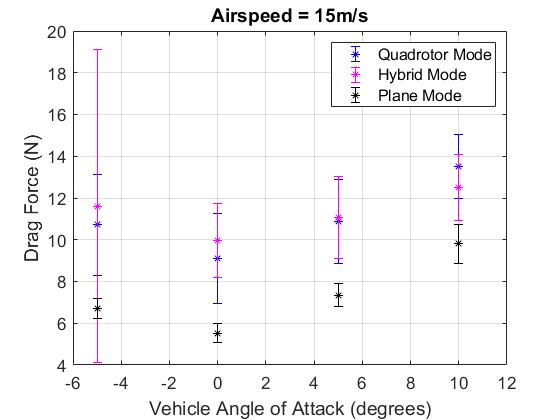}
    \end{tabular}
    \caption{Net Lift (left) and Drag (right) QuadPlane forces in different flight modes at airspeeds 5m/s, 11m/s and 15m/s. The highest lift and least drag is experimentally observed in Plane mode for both cruise airspeeds.}
    \label{fig: Lift and drag at 11, 15 and 5 m/s}
\end{figure}

Recorded $F_x,F_y,F_z$ data were sampled in Quad (vertical motors on), Plane (forward motor on), and Hybrid (all motors on) configurations to calculate lift and drag forces for each configuration. Sampling was done between $20s-30s$ for Quad mode ($10,000$ data points), $80s-90s$ for Hybrid mode ($10,000$ data points) and $110s-130s$ for Plane mode ($20,000$ data points) in each test sequence per Table \ref{tab: 5x7 test matrix}. These time stamps were selected to ensure flow had stabilized after changes made to control inputs.

Fig. \ref{fig: Lift and drag at 11, 15 and 5 m/s} shows experimental data for lift and drag at different airspeeds and angles of attack. Points denoted by stars show the data mean values; error bars represent one standard deviation from the mean values. Plots for $V_a \geq 11m/s$ (i.e. cruise conditions) indicate the vehicle has the least drag in Plane mode at all $\alpha_V$, as expected. Drag in Quadrotor mode is generally less than drag in Hybrid mode due to adverse flow interactions in Hybrid mode. Note that standard deviation error is lowest in Plane mode and highest in Hybrid mode consistent with raw data in Fig. \ref{fig: forces and moments - 11m/s aoa0} because flow interference which tends to have a periodic component is highest when all five rotors are spinning.

Lift and drag at $5m/s$, the transition phase, are shown in the top plots of Fig. \ref{fig: Lift and drag at 11, 15 and 5 m/s}. Note there was a load cell offset in $F_z$ due to poor calibration for the $\alpha_V = -5\degree$ at $V_a = 5m/s$ test case; this offset was corrected with a factor of $16.5N$ in $F_z$ based on other $\alpha_V$ trends at the same airspeed ($V_a = 5m/s$). 

Lift trends are similar for all tested airspeeds. Drag in Quad mode is almost the same as in Plane mode for low airspeed and low $\alpha_V$, while Hybrid mode has the highest drag. 
The simulation model used in \cite{QuadPlane_conference_paper}, based on ideal case assumptions of negligible flow interaction between rotors and the vehicle structure, always predicts higher lift and lower drag than were experimentally observed. Results show that flow interaction between structure and rotors plays a vital role in determining aerodynamic performance even for eVTOL control system studies.

\begin{figure}[!ht]
    \centering
    \begin{tabular}{cc}
        \includegraphics[width=0.47\linewidth]{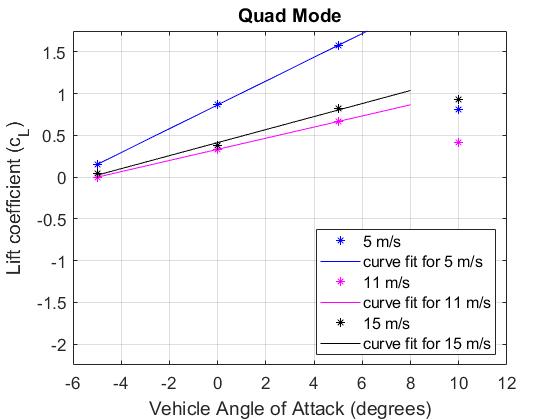} & \includegraphics[width=0.47\linewidth]{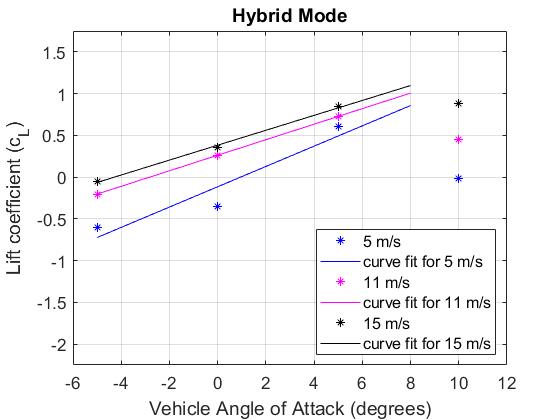} \\ 
        \includegraphics[width=0.47\linewidth]{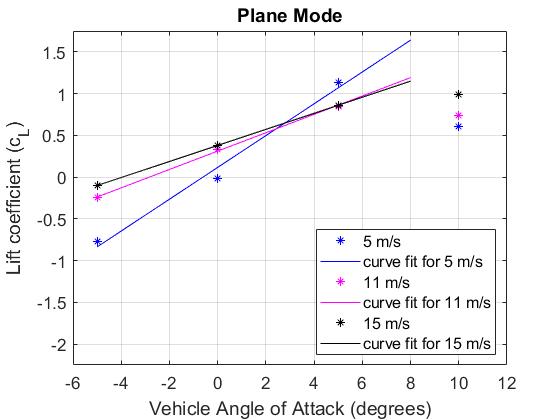} & 
        \includegraphics[width=0.47\linewidth]{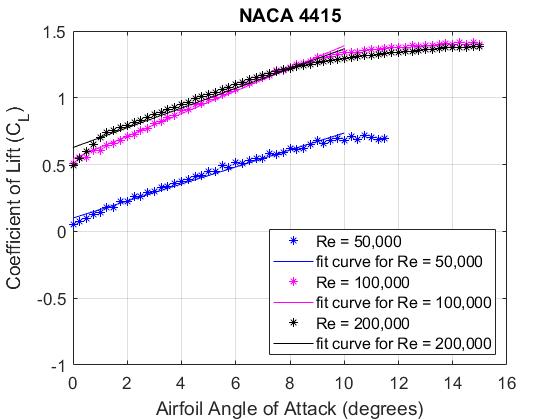}\\
        \\
    \end{tabular}
    \caption{Coefficients of lift obtained for Quad, Hybrid and Plane Modes for different airspeeds. Experimental data is shown with '$*$'; the fit curves are shown with solid lines. Curve fits are only shown up to wing stall angle.}
    \label{fig: cl for all modes vs AoA}
\end{figure}

Aerodynamic lift and drag coefficients were calculated from mean lift and drag forces per Eq. \ref{eq: QuadPlane Lift, Drag equations} for each flight mode at different $\alpha_V$ and $V_a$ values. Fig. \ref{fig: cl for all modes vs AoA} plots coefficient of lift ($C_{L_P}$) for the three flight modes as a function of $\alpha_V$. For comparison, NACA 4415 airfoil data \cite{naca_database} is plotted. Note that angle of attack for the wing $\alpha_{wing} = \alpha_V + 5\degree$ since the wing has an incidence angle of $5\degree$ relative to the QuadPlane body. Curve fits per Eq. \ref{eq:coeff of lift} are listed in Table \ref{tab - coefficient parameters for lift}. We assume a linear fit for coefficient of lift and only capture trends up to the stall angle as described in Section \ref{section: preliminaries}. 

\begin{table}[!ht]
    \centering
    \caption{Linear fit parameters obtained from curve fitting coefficient of lift ($C_{L_P}$) for different flight modes and Reynolds numbers. Post stall data is not considered for curve fitting. }
 
    \begin{tabular}{ c  c  c  c  c  c  c }
        \hline
        Flight Mode & Airspeed($m/s$) & Reynolds Number & $C_{L_{P_0}}$ &  $C_{L_{P_\alpha}}$& $R^2$ & RMSE\\
        \hline \hline
        \multirow{3}{*}{Quad} & 5 & 50,427 & 0.8652 & 0.1426 & 0.9999 & 0.01098\\
         & 11 & 110,939 & 0.3329 & 0.0667 & 1 & 0.00046\\
         & 15 & 151,281 & 0.4135 & 0.078 & 0.9923 & 0.04853\\
         \hline
        \multirow{3}{*}{Hybrid} & 5 & 50,427 & -0.1153 & 0.1216 & 0.898 & 0.2898\\
         & 11 & 110,939 & 0.2622 & 0.09302 & 1 & 0.002291\\
         & 15 & 151,281 & 0.3826 & 0.08929 & 0.9981 & 0.02689\\
        \hline
        \multirow{3}{*}{Plane} & 5 & 50,427 & 0.1165 & 0.1905 & 0.9843 & 0.1701\\
         & 11 & 110,939 & 0.3118 & 0.11 & 0.9989 & 0.02587\\
         & 15 & 151,281 & 0.3796 & 0.09621 & 1 & 0.001704\\
        \hline
        \multirow{3}{*}{NACA 4415 } & - & 50,000 & 0.1003 & 0.06363 & 0.9856 & 0.1457 \\
        & - & 100,000 & 0.5437 & 0.08457 & 0.9938 & 0.1263 \\
        & - & 200,000 & 0.628 & 0.07386 & 0.9691 & 0.2496 \\
        \hline
    \end{tabular}
    \label{tab - coefficient parameters for lift}
\end{table}

As expected, lift coefficient depends on Reynolds number and increases with Reynolds number for all flight modes. For low Reynolds numbers ($5m/s$ tunnel speed) wing stall occurs at approximately $\alpha_V = 5\degree$. However, the QuadPlane can still fly at this condition with vertical thrust assistance. For higher Reynolds numbers ($11 m/s$ and $15m/s$ tunnel speeds), stall angle is between $5\degree$ and $10\degree$. This is consistent with a NACA 4415 airfoil at similar Reynolds numbers \cite{naca_database}. For all flight modes, trends for cruise airspeeds of $11m/s$ and $15 m/s$ where Plane mode is dominant are similar.

\begin{figure}[!ht]
    \centering
        \includegraphics[width=0.47\linewidth]{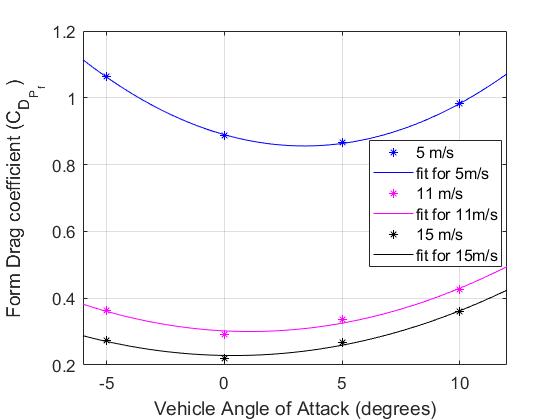}
    \caption{Form Drag Coefficient ($C_{D_{P_f}}$) at different airspeeds and angles of attack. Quadratic curve fits per Table \ref{tab - coefficient parameters for C_D_P_f} are also plotted.  $C_{D_{P_f}}$ is approximately constant across all QuadPlane flight modes.}
    \label{fig: form drag coeff}
\end{figure}

In Plane mode vertical propulsion modules are off so we set the Quad component of drag ($C_{D_Q}$) to zero. Using ($C_{L_P}$) values, we  compute $C_{D_{P_f}}$ using Eq. \ref{eq: coeff of drag - Plane mode} and \ref{eq: QuadPlane Lift, Drag equations} as follows: 

\begin{equation}
    C_{D_{P_f}} = \frac{2 D}{\rho S V_a^2} - \frac{C_{L_P}^2}{\pi e AR} 
    \label{eq: C_d_pf for plane mode}
\end{equation}
Note that Eq. \ref{eq: C_d_pf for plane mode} is only valid for Plane Mode. 
Per Section \ref{section: preliminaries}, $C_{D_{P_f}}$ represents form drag and is assumed to be the same over the three QuadPlane flight modes. $C_{D_{P_f}}$ is constant with respect to $\alpha_V$ for a standard aircraft. However, from Fig. \ref{fig: form drag coeff} for a QuadPlane it is a quadratic function of $\alpha_V$. The authors hypothesize that energy is being lost to the flow as a function of $\alpha_V$ due to the additional structure to accommodate the QuadPlane's quadrotor motors. We define $C_{D_{P_f}}$ as a quadratic function of $\alpha_V$ as given by Eq. \ref{eq: C_D_P_f general form}. Table \ref{tab - coefficient parameters for C_D_P_f} lists quadratic curve fit parameters for the data from Fig. \ref{fig: form drag coeff}.

\begin{equation}
    C_{D_{P_f}} = C_{D_{P_{f_0}}} + C_{D_{P_{f_\alpha}}}\alpha_V + C_{D_{P_{f_{\alpha^2}}}}\alpha_V^2 
    \label{eq: C_D_P_f general form}
\end{equation}

\begin{table}[!ht]
    \centering
    \caption{Form drag coefficient ($C_{D_{P_{f}}}$) quadratic curve fit values for different Reynolds numbers.}
    \begin{tabular}{  c   c   c   c   c   c   c   c  }
        \hline
        Airspeed($m/s$) & Reynolds Number (Re) & $C_{D_{P_{f_0}}}$ & $C_{D_{P_{f_\alpha}}}$ & $C_{D_{P_{f_{\alpha^2}}}}$ & $R^2$ & RMSE\\
        \hline \hline
         5 & 50,427  & 0.8896 & -0.01986 & 0.002921 & 0.9992 & 0.004529 \\
          11 & 110,939 & 0.3022 & -0.003535 & 0.001632 & 0.9678 & 0.0176\\
          15 & 151,281 & 0.2286 & -0.001117 & 0.001452 & 0.9805 & 0.01409\\
        \hline
    \end{tabular}
    \label{tab - coefficient parameters for C_D_P_f}
\end{table}

\begin{figure}[!ht]
    \centering
    \begin{tabular}{cc}
        \includegraphics[width=0.47\linewidth]{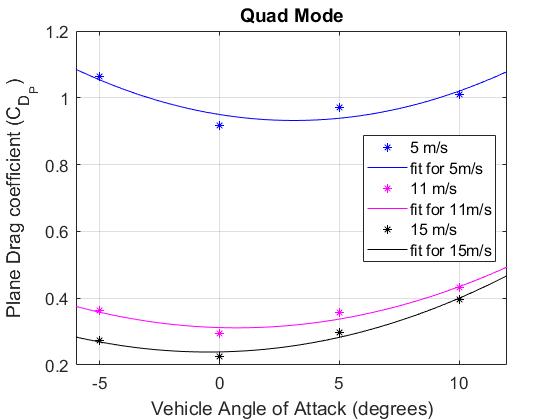} & 
        \includegraphics[width=0.47\linewidth]{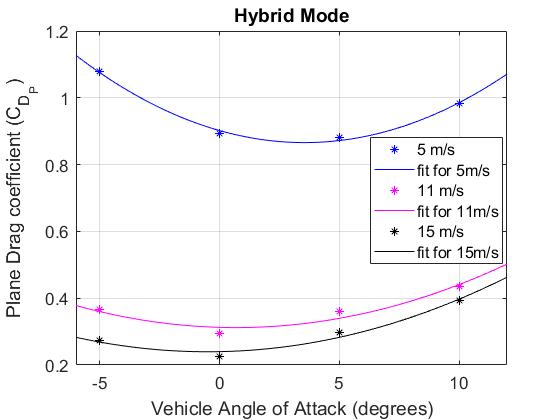} \\
        \multicolumn{2}{c}{\includegraphics[width=0.47\linewidth]{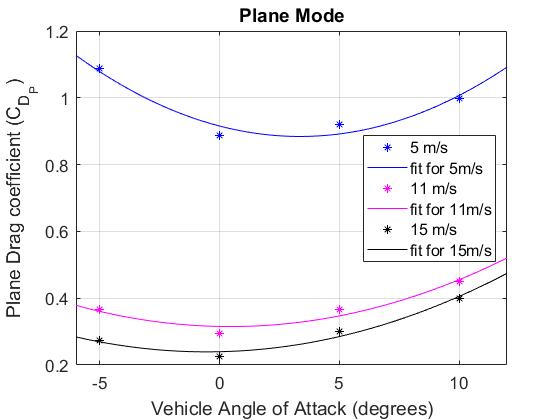}}
    \end{tabular}
    \caption{Plane Drag Coefficient ($C_{D_{P}}$) and corresponding curve fits for the QuadPlane at different airspeeds, angles of attack and flight modes (Quad - top left; Hybrid - top right; Plane - bottom).}
    \label{fig: Plane drag coeffs - all modes}
\end{figure}

Coefficient of drag for the QuadPlane's Plane component ($C_{D_P}$) for each flight mode uses $C_{L_P}$ values per Eq.\ref{eq: coeff of drag - Plane mode}. Fig. \ref{fig: Plane drag coeffs - all modes} plots $C_{D_P}$ for Quad, Plane and Hybrid modes. The lowest coefficient values occur at the highest airspeed. As expected, drag coefficients have a quadratic relationship with angle of attack and are modelled as shown in Eq. \ref{eq: C_D_P general form}. We notice that drag coefficients obtained for $V_a = 5m/s$ ($Re\approx 50,000$) are significantly higher than those seen at higher Reynolds numbers. This trend is consistent with drag coefficient data on a NACA 4415 airfoil at $Re = 50,000$ \cite{naca_database}. Quadratic curve fit coefficients are listed in Table \ref{tab - coefficient parameters for C_D_P}.

\begin{equation}
    C_{D_{P}} = C_{D_{P_{0}}} + C_{D_{P_{\alpha}}}\alpha_V + C_{D_{P_{{\alpha^2}}}}\alpha_V^2 
    \label{eq: C_D_P general form}
\end{equation}

\begin{table}[!ht]
    \centering
    \caption{Drag coefficients obtained from curve fitting on $C_{D_{P}}$ data for all flight modes.}
    \begin{tabular}{  c   c   c   c   c   c   c   c  }
        \hline
        Flight Mode & Airspeed($m/s$) & Re & $C_{D_{P_{0}}}$ & $C_{D_{P_{\alpha}}}$ & $C_{D_{P_{{\alpha^2}}}}$ & $R^2$ & RMSE\\
        \hline \hline
        \multirow{3}{*}{Quad} & 5 & 50,427 & 0.9499 & -0.01146 & 0.001848 & 0.8076 & 0.04669\\
         & 11 & 110,939 & 0.312 & -0.002021 & 0.001472 & 0.9332 & 0.02536\\
         & 15 & 151,281 & 0.2392 & 0.001396 & 0.001469 & 0.9708 & 0.0212\\
        \hline
        \multirow{3}{*}{Hybrid} & 5 & 50,427 & 0.9018 & -0.02029 & 0.002866 & 0.9911 & 0.01497\\
         & 11 & 110,939 & 0.3127 & -0.002008 & 0.001483 & 0.9161 & 0.02897\\
         & 15 & 151,281 & 0.24 & 0.001428 & 0.001429 & 0.9643 & 0.02301\\
        \hline
        \multirow{3}{*}{Plane} & 5 & 50,427  & 0.9154 & -0.01861 & 0.002777 & 0.9227 & 0.04301 \\
         & 11 & 110,939 & 0.3154 & -0.001331 & 0.001534 & 0.9221 & 0.03036\\
         & 15 & 151,281 & 0.2398 & 0.0016 & 0.001496 & 0.9708 & 0.02185\\
        \hline
    \end{tabular}
    \label{tab - coefficient parameters for C_D_P}
\end{table}

\begin{figure}[!ht]
    \centering
    \begin{tabular}{cc}
        \includegraphics[width=0.47\linewidth]{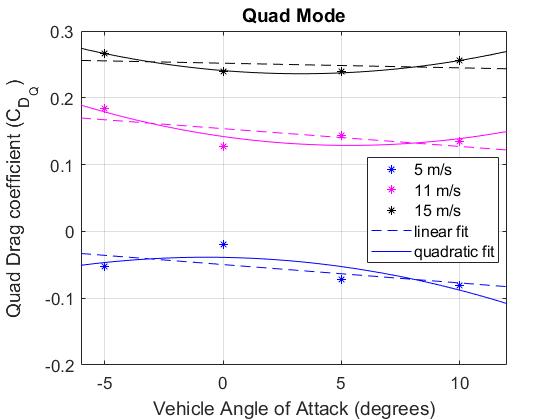} & 
        \includegraphics[width=0.47\linewidth]{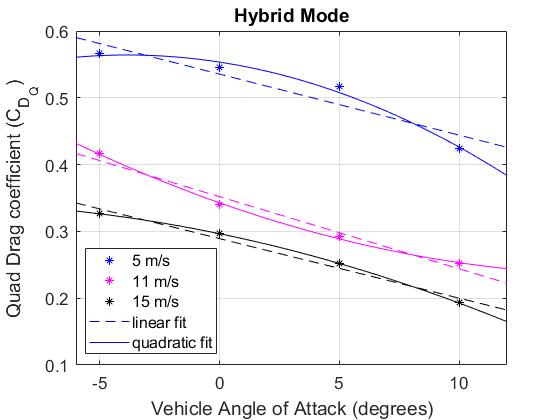} 
    \end{tabular}
    \caption{Quad Drag Coefficient ($C_{D_{Q}}$) linear and quadratic curve fits at different airspeeds and angles of attack.}
    \label{fig: Quad drag coeffs - all modes}
\end{figure}

Coefficient of drag for the QuadPlane's Quad component ($C_{D_Q}$), is computed for Quad and Hybrid modes using $C_{D_P}$ values per Eqs. \ref{eq: coeff of drag - Plane mode} and \ref{eq: QuadPlane Lift, Drag equations}. Linear and quadratic fits were considered for $C_{D_Q}$ as a function of $\alpha$: 
\begin{equation}
    C_{D_{Q}} = C_{D_{Q_{0}}} + C_{D_{Q_{\alpha}}}\alpha_V + C_{D_{Q_{\alpha^2}}}\alpha_V^2  
    \label{eq: C_D_Q general form}
\end{equation}
Fig. \ref{fig: Quad drag coeffs - all modes} overlays experimentally computed values with linear and quadratic fit curves. Curve fit parameters are listed in Table \ref{tab - coefficient parameters for C_D_Q}.

\begin{table}[!ht]
    \centering
    \caption{Curve fit drag coefficients $C_{D_{Q}}$ data for Quad and Hybrid modes at different Reynolds numbers.}
    \begin{tabular}{  c   c   c   c   c   c   c   c   c  }
        \hline
        Flight Mode & $V_a$($m/s$) & Re & Fit type & $C_{D_{Q_{0}}}$ & $C_{D_{Q_{\alpha}}}$ & $C_{D_{Q_{{\alpha^2}}}}$ & $R^2$ & RMSE\\
        \hline \hline
        \multirow{6}{*}{Quad} & \multirow{2}{*}{5} & \multirow{2}{*}{50,427} & Linear & -0.04963 & -0.002755 & - & 0.4285 & 0.03557 \\
        & & & Quadratic & -0.03888 & -0.0006044 & -0.0004301 & 0.6373 & 0.02833 \\
        \cline{2-9}
        & \multirow{2}{*}{11} & \multirow{2}{*}{110,939} & Linear & 0.1538 & -0.002662 & - & 0.4671 & 0.03179\\
        & & & Quadratic & 0.1421 & -0.005008 & 0.0004693 & 0.7575 & 0.0144 \\
        \cline{2-9}
        & \multirow{2}{*}{15} & \multirow{2}{*}{151,281} & Linear & 0.2518 & -0.0006953 & - & 0.11 & 0.02212 \\
        & & & Quadratic & 0.2408 & -0.002896 & 0.0004402 & 0.9914 & 0.002172 \\
        \hline
        \hline
        \multirow{6}{*}{Hybrid} & \multirow{2}{*}{5} & \multirow{2}{*}{50,427} & Linear & 0.5354 & -0.009128 & - & 0.8781 & 0.03803 \\
        & & & Quadratic & 0.5534 & -0.005538 & -0.000718 & 0.9867 & 0.01254 \\
        \cline{2-9}
        & \multirow{2}{*}{11} & \multirow{2}{*}{110,939} & Linear & 0.3519 & -0.01079 & - & 0.9758 & 0.01901\\
        & & & Quadratic & 0.3427 & -0.01264 & 0.0003697 & 0.9987 & 0.004444 \\
        \cline{2-9}
        & \multirow{2}{*}{15} & \multirow{2}{*}{151,281} & Linear & 0.2891 & -0.008903 & - & 0.9784 & 0.01479\\
        & & & Quadratic & 0.2965 & -0.007426 & -0.0002954 & 0.9999 & 0.0008046 \\
        \hline
    \end{tabular}
    \label{tab - coefficient parameters for C_D_Q}
\end{table}

\begin{figure}[!ht]
    \centering
    \begin{tabular}{cc}
        \includegraphics[width=0.47\linewidth]{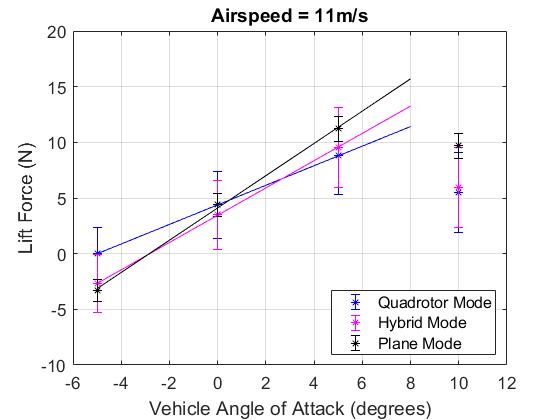}
        & \includegraphics[width=0.47\linewidth]{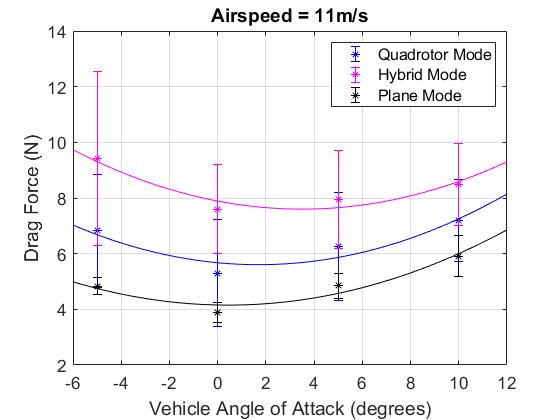}\\
    \end{tabular}
    \caption{Curve fits superimposed on Lift (left column) and Drag (right column) force data in different flight modes at $V_a = 11m/s$. All curve fits lie within one standard deviation of observed data points.}
    \label{fig: lift and drag with curve fits}
\end{figure}

Per Fig. \ref{fig: Quad drag coeffs - all modes}, both linear and quadratic fits reasonably represent observed data. $C_{D_Q}$ for Quad mode is almost constant, while linear fits describe Hybrid mode well. Hence, we consider the linear fit in our analysis.
Superimposing the above lift and drag curve fit models onto error bar data plots, confirms that derived model results are well within one standard deviation error bounds for all cases. Fig. \ref{fig: lift and drag with curve fits} illustrates this for $V_a = 11m/s$.

\begin{figure}[!ht]
    \centering
    \begin{tabular}{cc}
        \includegraphics[width=0.47\linewidth]{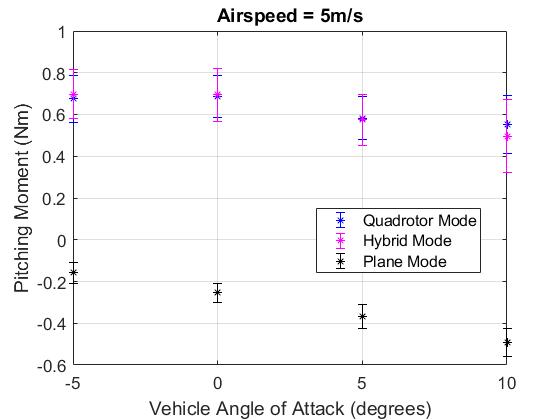} & 
        \includegraphics[width=0.47\linewidth]{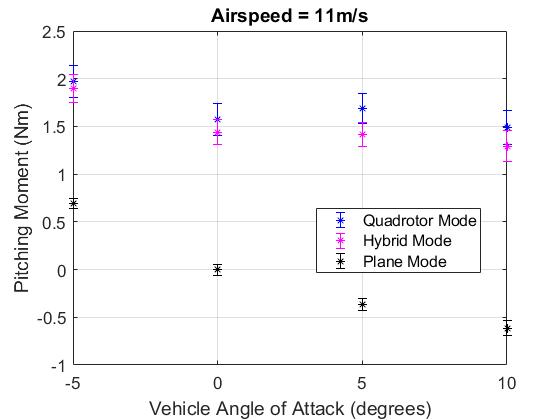} \\
    \multicolumn{2}{c}{\includegraphics[width=0.47\linewidth]{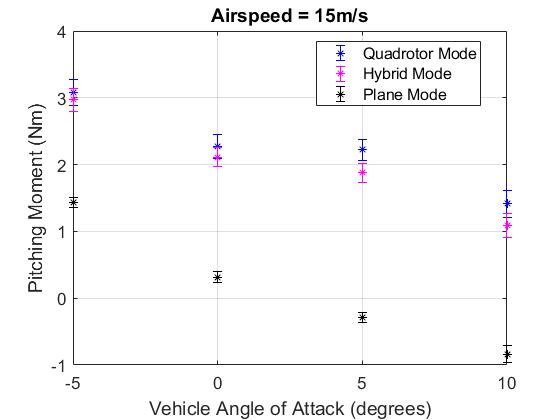}}   
    \end{tabular}
    \caption{Pitching Moment ($M_y$) as a function of vehicle angle of attack ($\alpha_V$) for the three test airspeeds with error bars denoting one standard deviation.}
    \label{fig: My error plot for all airspeeds}
\end{figure}

Next, we look at the pitching moment ($M_y$) in each flight condition. Fig. \ref{fig: My error plot for all airspeeds} shows the experimentally observed pitching moment in the three QuadPlane flight modes at different test airspeeds. Error bars represent one standard deviation from the mean values and the points denoted by stars show data mean values. Note that the reduced thrust from the rear vertical thrust motors creates a much larger pitching moment in both Quad and Hybrid modes than in Plane mode as discussed earlier in this section. Since the forward thrust vector acts slightly above the center of mass for the vehicle, we see a net negative pitching moment for $\alpha_V = 0\degree$ at low airspeeds ($V_a = 5m/s$). As expected \cite{stevens_aircraft_2015} pitch down moment tends to increase with an increase in vehicle angle of attack in all cases, with the most negative pitching moment observed at $\alpha_V = 10 \degree, V_a = 15m/s$ in Plane mode. Also, the standard deviation in Plane mode is much less than in Quad and Hybrid modes as the vehicle experiences much higher flow interactions and transient effects in Quad and Hybrid modes.

\begin{figure}[!ht]
    \centering
    \begin{tabular}{c}
        \includegraphics[width=0.47\linewidth]{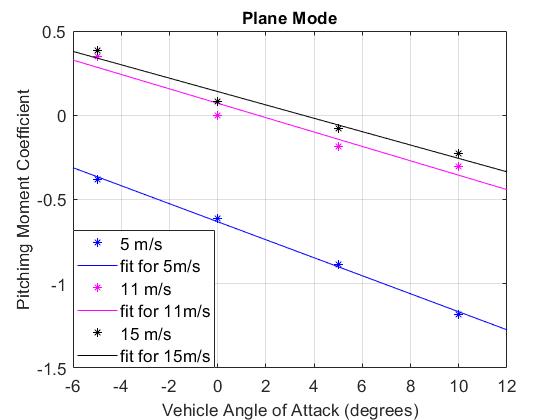}
    \end{tabular}
    \caption{Coefficient of Pitching Moment ($C_M$) as a function of vehicle angle of attack ($\alpha_V$) for Plane mode with neutral elevator position. Linear curve fits are overlayed.}
    \label{fig: cm - Plane mode}
\end{figure}

Coefficient of pitching moment for the vehicle structure ($C_M$) is calculated in the Plane mode per Eq. \ref{eq: pitching moment}. Since the vertical thrust motors are off and elevator is at a neutral position, Eq. \ref{eq: pitching moment} reduces to $ C_{M} = C_{M_{0}} + C_{M_{\alpha}} \alpha_V $, and we can calculate $C_{M_0}$ and $C_{M_\alpha}$. Fig. \ref{fig: cm - Plane mode} shows the net pitching moment coefficient values for Plane mode as a function of $\alpha_V$ at different airspeeds. Overlayed linear curve fits yield $C_{M_0}$ and $C_{M_\alpha}$ as listed in Table \ref{tab - coefficient parameters for C_M}.

\begin{table}[!ht]
    \centering
    \caption{$C_{M_0}$ and $C_{M_\alpha}$ parameters for the coefficient of moment ($C_M$) obtained in Plane mode.}
    \begin{tabular}{  c   c   c   c   c  c  }
        \hline
        Airspeed($m/s$) & Re & $C_{M_0}$ & $C_{M_{\alpha}}$ & $R^2$ & RMSE\\
        \hline \hline
         5 & 50,427 & -0.632 & -0.05345 & 0.9965 & 0.03558 \\
         11 & 110,939 & 0.0711 & -0.04272 & 0.9452 & 0.115\\
         15 & 151,281 & 0.1407 & -0.0397 & 0.9682 & 0.08051\\
        \hline
    \end{tabular}
    \label{tab - coefficient parameters for C_M}
\end{table}

\begin{figure}[!ht]
    \centering
    \begin{tabular}{cc}
        \includegraphics[width=0.47\linewidth]{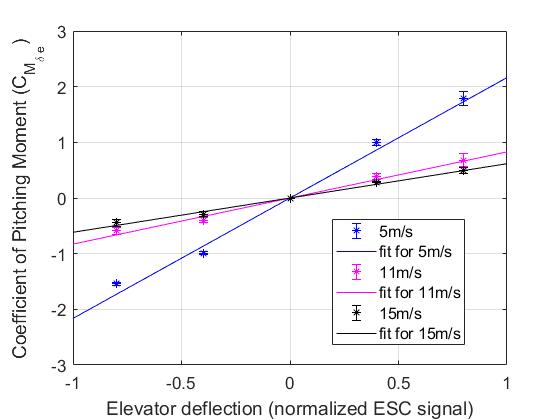}
    \end{tabular}
    \caption{$C_{M_{\delta e}}$ as a function of elevator deflection for different airspeeds with linear curve fits.}
    \label{fig: cm_e vs delta e}
\end{figure}

Next, we look at the contributions of the elevator deflection towards pitching moment. Pitching moment data was taken at time stamps consistent with the elevator deflection per Table \ref{tab: 5x7 test matrix}. The sampled data was averaged, and average nominal pitching moment from the vehicle structure was subtracted for each test case. The data was also averaged over $\alpha_V$ as $C_{M_\alpha}$ already accounts for variation with vehicle angle of attack. Fig. \ref{fig: cm_e vs delta e} shows the linear dependence of $C_{M_\delta e}$ with elevator deflection at different Reynolds numbers characterized by $V_a$. The coefficient parameters are tabulated in Table \ref{tab - coefficient parameters for C_M_e}.

\begin{table}[!ht]
    \centering
    \caption{Linear fit parameters for $C_{M_{\delta e}}$.}
    \begin{tabular}{  c   c   c   c  c  }
        \hline
    Airspeed($m/s$) & Re & $C_{M_{\delta e}}$ & $R^2$ & RMSE\\
        \hline \hline
    5 & 50,427 & 2.163 & 0.9921 & 0.2432 \\
    11 & 110,939 & 0.8286 & 0.9899 & 0.106\\
    15 & 151,281 & 0.616 & 0.9908 & 0.0749\\
    \hline
    \end{tabular}
    \label{tab - coefficient parameters for C_M_e}
\end{table}

\begin{figure}[!ht]
    \centering
    \begin{tabular}{cc}
        \includegraphics[width=0.47\linewidth]{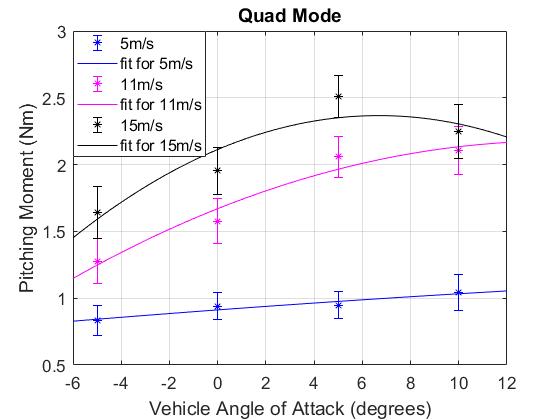} &
        \includegraphics[width=0.47\linewidth]{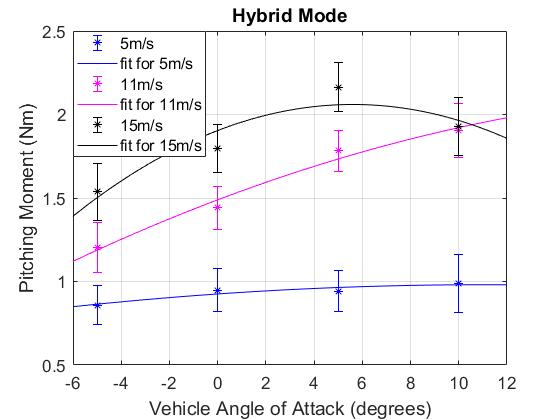}
    \end{tabular}
    \caption{Pitching moment from differential front and rear vertical thrust modules ($M_{\Delta T_{vert}}$) as a function of vehicle angle of attack ($\alpha_V$) with at different airspeeds ($V_a$).}
    \label{fig: moment from differential thrust}
\end{figure}

Fig. \ref{fig: moment from differential thrust} plots the pitching moment generated purely by the difference in thrust between the front and rear vertical propulsion modules ($M_{\Delta T_{vert}}$) as a function of vehicle angle of attack ($\alpha_v$) at the three test airspeeds. The data is plotted with mean values denoted by stars and one standard deviation denoted by error bars.  Pitching moment is higher for higher airspeeds as expected. We model $M_{\Delta T_{vert}}$ per Eq. \ref{eq: pitching moment from differential thrust} as a quadratic function of $\alpha_V$. Curve fits are overlayed on experimental data in Fig. \ref{fig: moment from differential thrust} and the quadratic fit coefficients for Quad and Hybrid modes are tabulated in Table \ref{tab - coeff parameters for differential thrust moment}. 

\begin{table}[!ht]
    \centering
    \caption{Quadratic fit parameters for $M_{\Delta T_{vert}}$.}
    \begin{tabular}{  c   c   c   c   c   c   c   c  }
        \hline
    Flight Mode & Airspeed($m/s$) & Re & $M_{\Delta T_{{vert}_0}}$ & $M_{\Delta T_{{vert}_\alpha}}$ & $M_{\Delta T_{{vert}_\alpha^2}}$ & $R^2$ & RMSE\\
    \hline \hline
    \multirow{3}{*}{Quad} & 5 & 50,427 & 0.9124 & 0.01333 & -0.0001248 & 0.9226 & 0.04117 \\
    & 11 & 110,939 & 1.67 & 0.07178 & -0.002539 & 0.9597 & 0.1379\\
    & 15 & 151,281 & 2.113 & 0.076 & -0.005691 & 0.8681 & 0.235\\
    \hline
    \multirow{3}{*}{Hybrid} & 5 & 50,427 & 0.9254 & 0.01001 & -0.0004523 & 0.8718 & 0.03434 \\
    & 11 & 110,939 & 1.489 & 0.05458 & -0.001133 & 0.9825 & 0.07329\\
    & 15 & 151,281 & 1.904 & 0.05568 & -0.004967 & 0.8777 & 0.1586\\
    \hline
    \end{tabular}
    \label{tab - coeff parameters for differential thrust moment}
\end{table}

So far,  we have listed all the longitudinal force and moment coefficients. For the lateral moments, we consider the rolling and yawing moment as a function of aileron and rudder deflection each. We also consider the side force as a function of rudder deflections. The test matrix did not include testing for sideslip and hence side force and yawing moment as a function of sideslip is not included in this work.

\begin{figure}[!ht]
    \centering
    \begin{tabular}{cc}
        \includegraphics[width=0.47\linewidth]{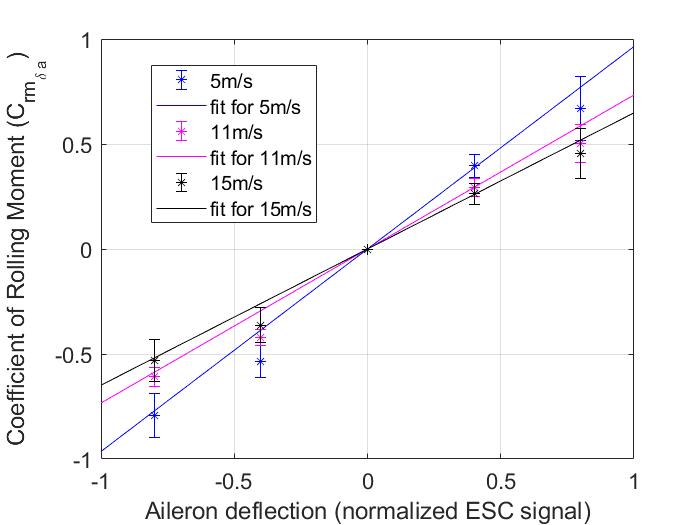} &
        \includegraphics[width=0.47\linewidth]{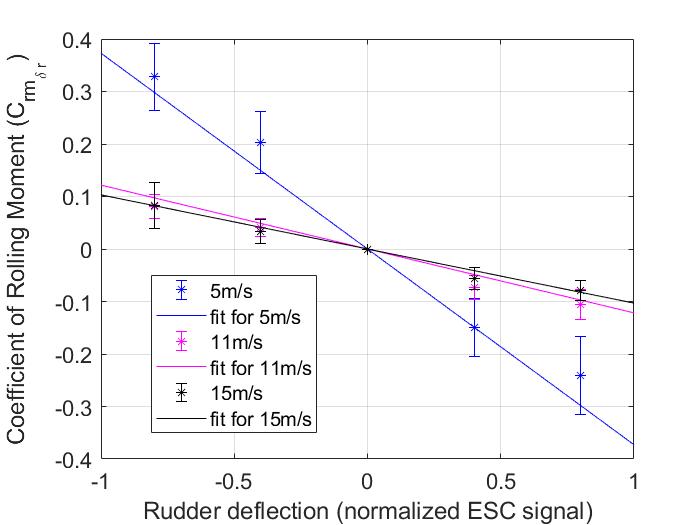}
    \end{tabular}
    \caption{Coefficient of rolling moment $C_{rm}$ as a function of aileron $\delta a$ and rudder deflection $\delta r$.}
    \label{fig: c_rm vs dela and delr}
\end{figure}

\begin{figure}[!ht]
    \centering
    \begin{tabular}{cc}
        \includegraphics[width=0.47\linewidth]{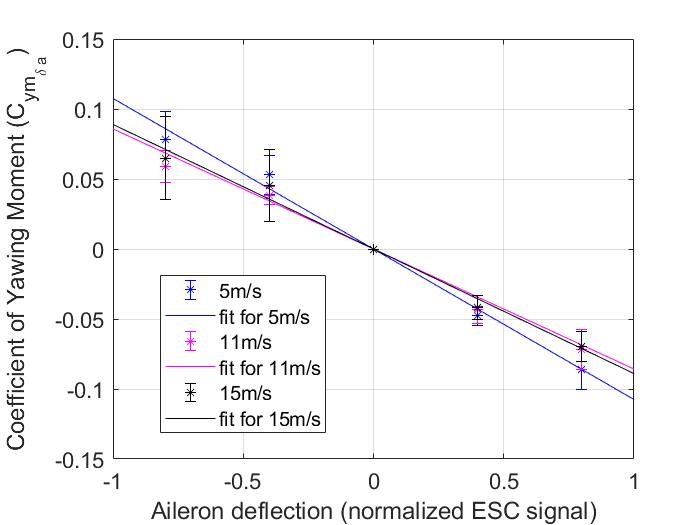} &
        \includegraphics[width=0.47\linewidth]{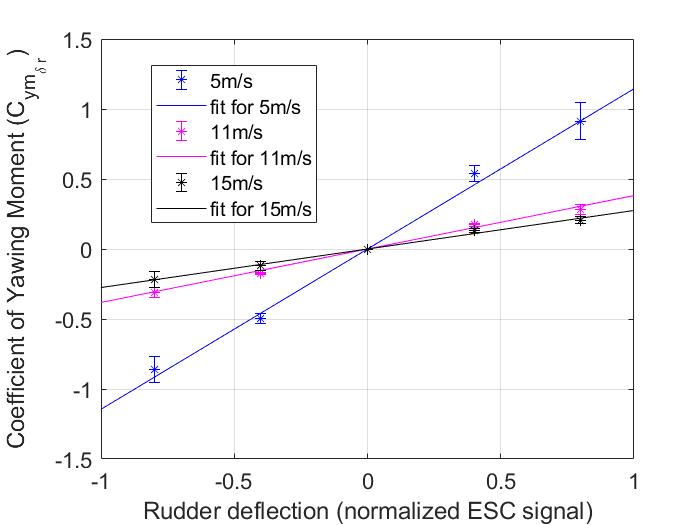}\\
    \end{tabular}
    \caption{Coefficient of yawing moment $C_{ym}$ as a function of aileron $\delta a$ and rudder deflection $\delta r$.}
    \label{fig: c_ym vs dela and delr}
\end{figure}

As seen in Fig. \ref{fig: forces and moments - 11m/s aoa0}, aileron and rudder deflections occur between $t = 130-170s$ and $t= 210-250s$ respectively, causing the vehicle to experience rolling and yawing moments. As in the case of elevator deflection's contribution towards pitching moment, rolling and yawing moment data was averaged over all angles of attack for a given airspeed. Rolling and yawing moments were modelled as linear function of both aileron and rudder deflections per Eq. \ref{eq: rolling and yawing moment and coeffs}. Fig. \ref{fig: c_rm vs dela and delr} and Fig. \ref{fig: c_ym vs dela and delr} show the linear trends for all coefficients while curve fit parameters are listed in Table \ref{tab - coefficient parameters for C_rm} and Table \ref{tab - coefficient parameters for C_ym}. 

\begin{table}[!ht]
    \centering
    \caption{Linear fit parameters for coefficient of rolling moment $C_{rm}$ as a function of $\delta a$ and $\delta r$.}
    \begin{tabular}{  c   c   c   c   c   c   c   c  }
        \hline
    Airspeed($m/s$) & Re & $C_{rm_{\delta a}}$ & $R^2$ & RMSE & $C_{rm_{\delta r}}$ & $R^2$ & RMSE\\
        \hline \hline
    5 & 50,427 & 0.9642 & 0.9778 & 0.0914 & -0.3728 & 0.9689 & 0.0419\\
    11 & 110,939 & 0.7336 & 0.9721 & 0.0781 & -0.1217 & 0.9585 & 0.0158\\
    15 & 151,281 & 0.6484 & 0.9777 & 0.0616 & -0.1031 & 0.9830 & 0.0086\\
    \hline
    \end{tabular}
    \label{tab - coefficient parameters for C_rm}
\end{table}

\begin{table}[!ht]
    \centering
    \caption{Linear fit parameters for coefficient of yawing moment $C_{ym}$ as a function of $\delta a$ and $\delta r$.}
    \begin{tabular}{  c   c   c   c   c   c   c   c  }
        \hline
    Airspeed($m/s$) & Re & $C_{ym_{\delta a}}$ & $R^2$ & RMSE & $C_{ym_{\delta r}}$ & $R^2$ & RMSE\\
        \hline \hline
    5 & 50,427 & -0.1075 & 0.9903 & 0.0067 & 1.145 & 0.9950 & 0.0514\\
    11 & 110,939 & -0.08574 & 0.9836 & 0.0070 & 0.3819 & 0.9941 & 0.0186\\
    15 & 151,281 & -0.08898 & 0.9864 & 0.0066 & 0.2752 & 0.9936 & 0.0139\\
    \hline
    \end{tabular}
    \label{tab - coefficient parameters for C_ym}
\end{table}

\begin{figure}[!ht]
    \centering
    \begin{tabular}{cc}
        \includegraphics[width=0.47\linewidth]{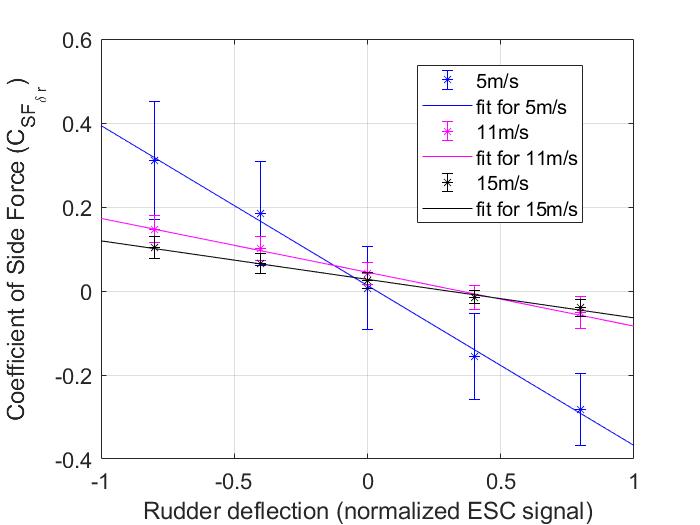}
    \end{tabular}
    \caption{Coefficient of side force $C_{SF}$ as a function of rudder deflection.}
    \label{fig: c_sf}
\end{figure}

Finally, we analyze the side force generated by rudder deflections. The coefficient of side force varies as a linear function of the rudder deflection $\delta r$ as described in Eq. \ref{eq; side force equation}. Again, the side force values were averaged over the vehicle angles of attack, same as for all control surface deflection based coefficients. Fig. \ref{fig: c_sf} shows the linear curves capturing side-force data, while the fit coefficients are tabulated in Table \ref{tab - coefficient parameters for C_SF}. 

\begin{table}[!ht]
    \centering
    \caption{Linear fit parameters for coefficient of yawing moment $C_{ym}$ as a function of $\delta a$ and $\delta r$.}
    \begin{tabular}{c   c   c   c   c   c}
        \hline
    Airspeed($m/s$) & Re & $C_{SF_0}$ & $C_{SF_{\delta r}}$ & $R^2$ & RMSE\\
        \hline \hline
    5 & 50,427 & 0.0130 & -0.3808 & 0.9966 & 0.0163\\
    11 & 110,939 & 0.0446 & -0.1283 & 0.9946 & 0.0069\\
    15 & 151,281 & 0.0277 & -0.0917 & 0.9943 & 0.0051\\
    \hline
    \end{tabular}
    \label{tab - coefficient parameters for C_SF}
\end{table}

Note that side force is nearly zero in all cases for zero rudder deflections, as seen in Fig. \ref{fig: c_sf} and the values of $C_{SF_0}$ in Table \ref{tab - coefficient parameters for C_SF}. Small inaccuracies occur possibly due to the vehicle experiencing a small sideslip angle for some of the tests. 

Thus far we have tabulated aerodynamic coefficients as a function of $\alpha_V$ at three distinct Reynolds numbers. The next step is to perform two-dimensional interpolation enabling aerodynamic coefficient estimation as a function of measured airspeed as well as $\alpha_V$. Fig. \ref{fig: 3D plot Cl_P, CD_P and CD_Q for hybrid mode} shows 3D plots for the aerodynamic force coefficients $C_L$/$C_{D_P}$/$C_{D_Q}$ in Hybrid mode as a function of $\alpha_V$ and $V_a$. Note that Hybrid mode best motivates this interpolation procedure since the QuadPlane typically accelerates or decelerates while in Hybrid mode.  Each triangle in the illustrated triangular mesh is numbered, and its planar equation is stored in a lookup table. We can rapidly index $\alpha_V$ and $V_a$ into this lookup table, obtain the three $C_L$/$C_{D_P}$/$C_{D_Q}$ planar equations, then compute interpolated coefficients for a specific ($\alpha_V$, $V_a$).

\begin{figure}[!ht]
    \centering
    \begin{tabular}{cc}
        \includegraphics[width=0.47\linewidth]{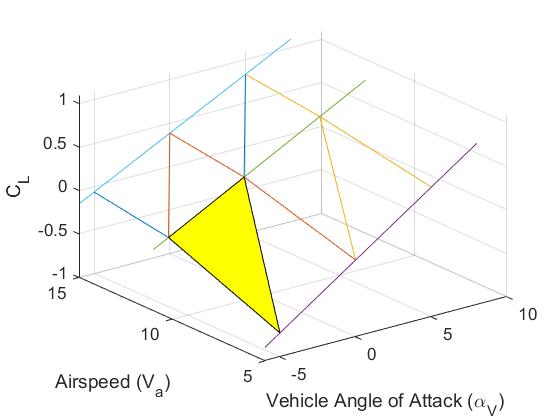}
        &
        \includegraphics[width=0.47\linewidth]{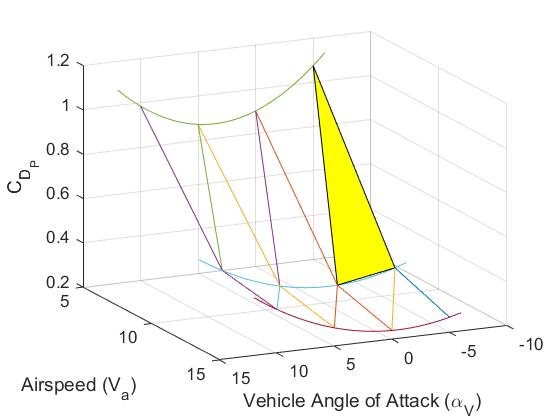}\\  
        \multicolumn{2}{c}{\includegraphics[width=0.47\linewidth]{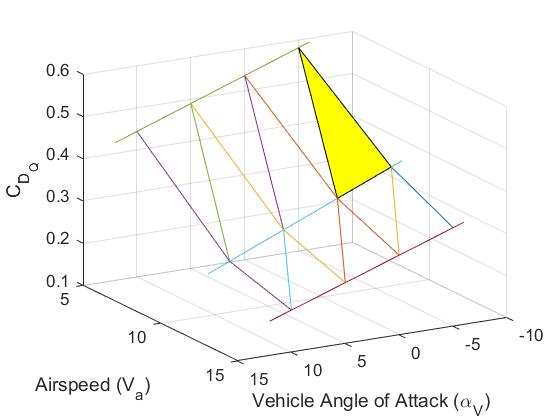}} 
    \end{tabular}
    \caption{Three dimensional plots of $C_{L}$ (top-left), $C_{D_P}$ (top-right) and $C_{D_Q}$ (bottom) in Hybrid mode for different Reynolds numbers (characterized by $V_a$) and $\alpha_V$.}
    \label{fig: 3D plot Cl_P, CD_P and CD_Q for hybrid mode}
\end{figure}

Let's consider that at some point during transition from Quad to Plane mode, $V_a = 9m/s$ and $\alpha_V = -4\degree$. In order to get the aerodynamic force coefficients for these conditions, we need the equation of the plane containing this point. Using a simple algorithm, it can be seen that the triangles highlighted as yellow in Fig. \ref{fig: 3D plot Cl_P, CD_P and CD_Q for hybrid mode} would contain this point. We get the corresponding planar equations for the triangles from the lookup table as follows:

\begin{equation}
\begin{aligned}
    0.0923 \alpha_V + 0.0860 V_a - 0.9920 C_{L} - 0.6863 & = 0\\ 
    0.0094 \alpha_V + 0.1183 V_a + 0.9929 C_{D_P} - 1.6122 & = 0 \\
    0.0108 \alpha_V + 0.0292 V_a + 0.9995 C_{D_Q} - 0.6728 & = 0
    .
\end{aligned}
\end{equation}

Solving for the aerodynamic force coefficients, we obtain $C_L = -0.2838$, $ C_{D_P} = 0.5893$, and  $C_{D_Q} = 0.4534$
for $V_a = 9m/s$ and $\alpha_V = -4\degree$.

\section{Discussion} \label{Section: Discussion}

This paper has presented a simple QuadPlane aircraft design to support experimental aerodynamic force and moment modelling for a hybrid eVTOL vehicle using polynomial curve fits. Wind tunnel tests on the airframe reveal high drag caused by flow interactions. The vertical thrust modules disturb oncoming flow and cause the angle of attack to drastically vary over the wingspan. In future work, it would be advantageous to combine Computational Fluid Dynamics (CFD) analysis with experimental characterization to maximize understanding of flow behaviors. With CFD, special areas of interest can be investigated, such as the impact of propeller downwash on wing lift, drag and pitching moment. 

As discussed in Section \ref{section: full vehicle windtunnel tests}, increased pitching moment is observed when vertical motors are spinning at appreciable forward velocities. The tests conducted for this analysis provided equal input commands to all four vertical motors. However, a controller would reduce input to the front two vertical motors and increase input to the rear ones to counter the positive pitching moment experienced. QuadPlane control laws can also mix elevator deflections and motor inputs to obtain the desired pitching moment. As seen from $200s - 210s$ in Fig.\ref{fig: forces and moments - 11m/s aoa0}, $80\%$ deflection of the elevator itself creates sufficient negative pitching moment to counter the effects of the spinning vertical motors. Alternatively, to counter the moment, the front motors each need to produce $0.88N$ less thrust and rear motors $0.88N$ more thrust to balance pitching moment. For the tested QuadPlane, $0.88N$ is $15\%$ of total hover thrust needed to balance weight. 

Even though flow interactions will always exist on designs like the QuadPlane, specific aspects can be improved in a future upgrade. The transverse beams used to mount the vertical propulsion modules in the current design further disturb oncoming airflow for the wings. The use of longitudinally-aligned beams with a twin boom tail design would reduce upstream flow disturbance. The use of multiple smaller vertical propulsion units spread over the wingspan would further reduce upstream flow disturbance but would add complexity.   

Due to cost considerations, it is not often possible to use large wind tunnels to conduct dynamic thrust tests. Smaller wind tunnels are inexpensive and can be used for such tests as shown in this paper. However, performing dynamic thrust tests at $\alpha_p$ close to $90\degree$ in the $2' \times 2'$ wind tunnel was challenging due to the small size of the test section. At large $\alpha_p$ values, the wind tunnel walls were too close downstream of the propeller to yield accurate free stream thrust results, a similar challenge to that described in \cite{wt_techniques_tiltwing}.    

In this work, the same propulsion module was used for both vertical and forward thrust as the static thrust values for the module was found to be sufficient for sustaining forward flight. However, the observed dynamic thrust for the forward propulsion module at higher airspeeds was lower than expected - only $44\%$ of its static counterpart at $V_a = 15m/s$. As a consequence, we can observe in Fig. \ref{fig: forces and moments - 11m/s aoa0} that there is insufficient thrust for the vehicle to reach $V_a \geq 11m/s$. A forward thrust propulsion module supporting a larger pitch propeller and higher torque motor is required to sustain steady level flight at $V_a \geq 15m/s$ with the current vehicle configuration.


\section{Conclusion} \label{Section: Conclusion}
This paper has presented the experimental aerodynamic analysis of a hybrid eVTOL QuadPlane prototype capable of hovering as a quadrotor, cruising as an aircraft, and executing transitions between these two modes. Propulsion module dynamic thrust was characterized over a wide sweep of propeller angles of attack. Wind tunnel tests were conducted to observe and record aerodynamic forces and moments in all three flight modes. Aerodynamic coefficients of lift, drag, side-force, pitching, rolling and yawing moments were computed and compared. Lift and drag curves generated using coefficient-based models were confirmed to accurately represent experimental data. 

The QuadPlane exhibits high drag and low dynamic thrust due to flow interactions. The aircraft naturally pitches up when vertical motors spin in appreciable free stream flow conditions.  This behavior can be attenuated with elevator deflection or can be leveraged to facilitate transition to hover mode during deceleration periods. 
Coefficient based models developed in this work can serve as a baseline for developing hybrid control schemes for eVTOL platforms including transition periods. The tests and analyses described in this paper can also inform aerodynamic models for future hybrid eVTOL designs.

\section*{Acknowledgements}
We would like to thank Christopher Chartier, Sidharth Anantha, Prashin Sharma, Prince Kuevor, and Matthew Romano for their assistance with instrumentation integration and wind tunnel testing support. This research was supported in part by a grant from the Kahn Foundation.

\bibliography{references}

\end{document}